\documentclass{article}

\PassOptionsToPackage{numbers,sort&compress}{natbib}
\usepackage[preprint]{neurips_2025}

\makeatletter
\renewcommand{\@notice}{\enlargethispage{2\baselineskip}}
\makeatother

\usepackage[utf8]{inputenc}
\usepackage[T1]{fontenc}
\usepackage{amsmath,amssymb}
\usepackage{graphicx}
\usepackage{float}
\usepackage{booktabs}
\usepackage{multirow}
\usepackage{siunitx}
\usepackage{caption}
\usepackage{subcaption}
\usepackage[protrusion=true,expansion=false]{microtype}
\usepackage{hyperref}
\usepackage{url}

\hypersetup{
  hidelinks,
  pdftitle={FISICA: A Deployed Service for Plantar-Pressure and Photographic
            Posture Assessment with Ontology-Grounded Recommendation},
  pdfauthor={Juhwan Song, Heejung Kim, Juntae Noh, Jonghak Ryu, Huiju Park,
             Junseong Lee, Dohyeon Ahn, Byungwoo Jo}
}
\DeclareCaptionLabelSeparator{bar}{~|~}
\title{FISICA: A Deployed Service for Plantar-Pressure and\\
Posture Assessment with Ontology-Grounded Recommendation}

\author{%
  Juhwan Song\textsuperscript{1} \quad Heejung Kim\textsuperscript{2} \quad
  Juntae Noh\textsuperscript{2} \quad Jonghak Ryu\textsuperscript{2} \quad
  Huiju Park\textsuperscript{2} \\
  \bfseries Junseong Lee\textsuperscript{2} \quad
  Dohyeon Ahn\textsuperscript{2} \quad Byungwoo Jo\textsuperscript{2} \\[0.45em]
  \normalfont\normalsize\textsuperscript{1}Department of Computer Science,
  University of Illinois Urbana-Champaign \\
  \normalfont\normalsize\textsuperscript{2}Care\&Co., Busan, South Korea \\[0.2em]
  \normalfont\small\texttt{juhwans3@illinois.edu} \\
  \normalfont\small\texttt{\{heejung,\,juntae,\,jonghak,\,huiju,\,seong,\,dohyeon,\,carencoinc\}@carenco.kr}
}

\begin{document}
\maketitle

\begin{abstract}
FISICA is a body-assessment and recommendation service running in production.
One standing session with two photographs returns foot-loading measures,
posture coordinates, a driven 3D avatar, a visual report, and ranked shoe and
exercise candidates. Measurement comes from a purpose-built scale carrying
\num{634} force-sensitive elements on a \SI{1}{\centi\metre} grid and four load
cells, and a rule-based evaluator controls every recommendation while a
language model only explains the stored result. The method contribution is the
avatar. Instead of mapping a measured angle onto a rig through a tuned gain, we
measure the avatar with the same function used on the subject and solve until
the two agree, on a sampling-invariant spinal metric that separated a normal
from a kyphotic record by \SI{7.2}{\degree} against \SI{0.9}{\degree} for a
single-joint formulation. In production, general APIs respond at a
\SI{0.023}{\second} median, plantar-pressure analysis at \SI{0.45}{\second},
and recommendation at \SIrange{2.16}{2.26}{\second} with the rule-based portion
under one second in every trial. The served keypoint graph reaches \num{0.960}
PCK@0.2 on public data, and the catalog holds \num{699} shoes with \num{10500}
typed facts. An approved study supplies the radiographic reference for the
validation still ahead.
\end{abstract}

% ==========================================================================
\section{Introduction}
% ==========================================================================

Standing-posture and foot-loading tests usually need radiography or a
laboratory \cite{fortin2011clinical,singla2014postural}. Both are costly and
hard to reach outside a clinic. FISICA replaces that entry step with a
pressure-sensing scale and two RGB photographs.

\paragraph{What the service delivers.} One standing session produces six
outputs: a plantar-pressure map with regional loads and foot dimensions, a
foot-type label, posture coordinates with alignment indicators, a driven 3D
avatar, a ranked list of shoes with the reason for each match, and a list of
exercises under safety rules. Every output is stored with the measurement
record and with the model, rule, and catalog versions that produced it.

\paragraph{What is new.} The measurement components are established: a
resistive pressure array, a keypoint detector, a rule evaluator. The
contribution is what happens after measurement. A system that shows a body
model next to a number usually drives the model from the number through a
hand-tuned mapping, so the figure illustrates the finding and nothing checks
that the two agree. We instead measure the avatar with the same function
applied to the subject and solve until the two match. Doing so requires a
spinal metric whose value does not change with how densely the spine is
sampled, because the subject chain has five points and the avatar chain has
nineteen. The metric we adopt for that reason also turns out to discriminate:
on two records it separates a normal from a kyphotic posture by
\SI{7.2}{\degree}, where the single-joint formulation it replaced separates
them by \SI{0.9}{\degree}. Alongside this we report two things that are rarely
published for a service of this kind: derived-angle error as a function of
subject scale rather than averaged over it, and the service-level cost of every
feature in production.

\paragraph{Contributions.}
\begin{enumerate}
  \item A service architecture that links pressure and RGB capture to storage,
  analysis, recommendation, visualization, and reporting, with version
  provenance on every result (Section~\ref{sec:arch}).

  \item A plantar-pressure pipeline that converts a \num{645}-byte frame into
  foot dimensions, regional loads, and a nine-class label
  (Section~\ref{sec:pressure}).

  \item A measured account of the deployed vision graph on public data. It
  separates the accuracy cost of two deployment choices and reports derived
  angle error as a function of subject scale (Section~\ref{sec:vision-eval}).

  \item A visualization method in which the avatar is measured with the same
  function used on the subject and solved until the two agree, resting on a
  spinal metric invariant to sampling density. It separates a normal from a
  kyphotic record by \SI{7.2}{\degree}, against \SI{0.9}{\degree} for the
  single-joint formulation it replaced (Section~\ref{sec:avatar}).

  \item A recommendation layer in which numeric observations activate needs,
  and rules filter and rank a fixed catalog before a language model explains
  the stored result (Section~\ref{sec:recommendation}).

  \item A service-level performance account from the production and
  development clusters, split by feature and by pipeline stage. The rule-based
  portion of every recommendation held under one second in every trial, and
  the end-to-end figures match or beat published baselines measured at a looser
  layer (Section~\ref{sec:perf}).
\end{enumerate}

\paragraph{Scope.} We describe a deployed system and report what we measured on
it. Recommendation scores are catalog ranks, not probabilities of benefit. The
class labels and posture indicators are analysis outputs, not diagnoses. FISICA
is a fitness and wellness service.

% ==========================================================================
\section{Related work}
% ==========================================================================

Human pose estimation moved from part-based models
\cite{yang2013articulated} to direct regression \cite{toshev2014deeppose} and
then to top-down detectors that find people first and regress keypoints per
person \cite{he2017maskrcnn,ren2015fasterrcnn,xiao2018simple,sun2019hrnet},
built on convolutional backbones with multi-scale features
\cite{he2016resnet,lin2017fpn} and more recently on transformers
\cite{xu2022vitpose}. On-device pipelines trade accuracy for speed
\cite{cao2019openpose,bazarevsky2020blazepose,lugaresi2019mediapipe,jiang2023rtmpose}.
We use such a model as a component and measure the exported graph. Our
evaluation follows the COCO keypoint protocol
\cite{lin2014coco,ronchi2017benchmarking,andriluka2014mpii}.

Photographic posture measures and the craniovertebral angle are established,
but they depend on camera placement, landmark definition, and rater technique
\cite{fortin2011clinical,singla2014postural,yip2008cva,raine1997posture,ferreira2010photogrammetry}.
Section~\ref{sec:angles} gives the model-side counterpart to that reliability
question.

Systems that present a posture finding on a body model normally drive the model
from the measured value through a hand-tuned mapping. The figure then
illustrates the finding, and nothing in the system checks that the figure and
the number agree. We are not aware of a posture system that measures its own
avatar with the function applied to the subject and solves until the two match.
Section~\ref{sec:avatar} gives that construction, together with the property a
metric must have for the comparison to be well posed at all.

Pressure platforms measure load distribution and footprint geometry. The arch
index and its anatomical thirds are the standard reduction
\cite{cavanagh1987archindex,razak2012plantar,orlin2000plantar,hughes1991reliability,wafai2015plantar}.
We adopt that reduction directly.

The gap between a model and a running service covers versioning, provenance,
and technical debt \cite{sculley2015debt,paleyes2022challenges,amershi2019se4ml}.
Our service boundary, in which a language model may only explain a decision
made by a rule evaluator, answers that concern.

% ==========================================================================
\section{System architecture}
\label{sec:arch}
% ==========================================================================

The service has seven parts: a pressure-sensing scale, frontal and lateral RGB
capture, a measurement service, learned and geometric analysis components, a
versioned domain graph, footwear and exercise evaluators, and an avatar client
with a conversational interface. We build and operate all seven. The sensor
board and firmware, the GPU cluster, the microservices, the knowledge base,
the catalog, and the 3D client are one system under one team, which is why the
provenance chain in this section reaches from a raw sensor frame to a ranked
product without an external boundary in between.

A measurement record holds capture time, source, application and firmware
provenance, and the pressure or vision payload. Analysis services add derived
observations without overwriting the source. Recommendation services receive a
compact projection of those observations, not the raw image or the full
pressure matrix. This boundary limits exposure of dense data and makes each
answer reproducible from the stored versions.

\begin{figure}[t]
\centering
\includegraphics[width=\textwidth]{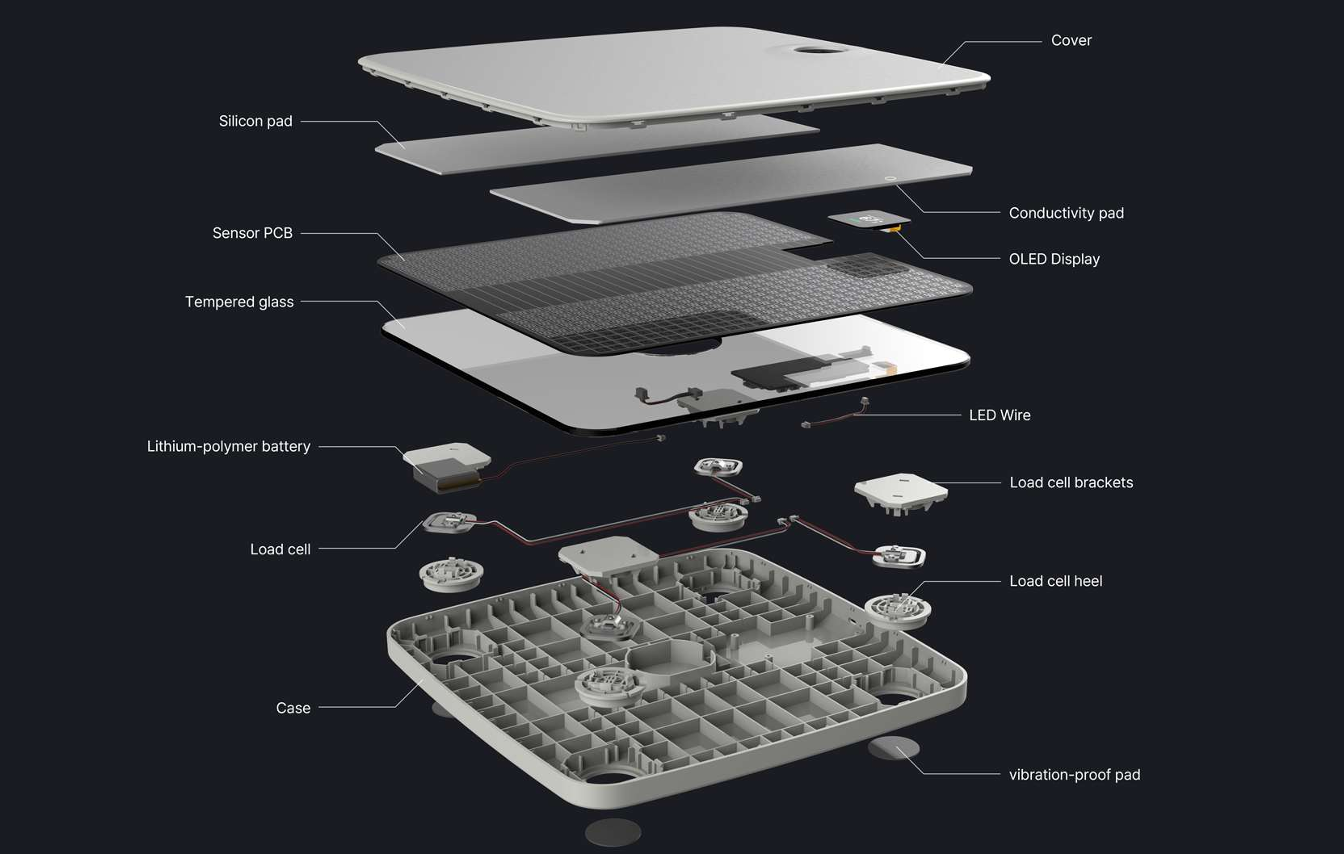}
\caption{Exploded view of the FISICA Scale. The device integrates the cover,
silicone and conductive layers, a pressure-sensor PCB, tempered glass, four
load-cell supports, a main controller, an OLED display, a rechargeable
battery, and the enclosure. The pressure array measures spatial loading. The
load cells provide total body mass.}
\label{fig:hardware}
\end{figure}

We describe the device first, in Figure~\ref{fig:hardware} and
Section~\ref{sec:hw}, because the sensing pitch it fixes is what later lets
foot dimensions be reported in millimetres without a camera reference.

\subsection{FISICA Scale hardware}
\label{sec:hw}

The scale is a self-contained standing platform. It needs no camera
calibration target and no external reference object, because the sensing pitch
is a physical constant of the board. Table~\ref{tab:hardware} gives the
specification.

\paragraph{Layer stack.} Load reaches the sensors through a fixed stack. A
tempered glass top of $300\times300\times5$~mm spreads the load and gives the
standing surface. Below it an ABS and TPR cover encloses two silicone layers,
one of them carrying a conductive pad, which set the compliance between glass
and sensor and keep the response repeatable across the array. The sensor PCB
sits under those layers. Four load cells are mounted on ABS brackets and heels
at the corners of the enclosure, outside the pressure array, on
vibration-damping pads that isolate the cells from cabinet resonance. The main
PCB, the 1.3-inch OLED, and the \SI{1200}{\milli\ampere\hour} lithium-polymer
cell occupy the remaining volume.

\paragraph{Two independent measurement paths.} The pressure path and the mass
path never share a converter (Table~\ref{tab:paths}). Keeping them separate is
what lets the spatial distribution and the absolute mass be trusted
independently: a drift in one path does not propagate into the other, and the
per-cell sum can be checked against the array sum as a validity test. The
ESP32-S3 on the main PCB handles the display, the battery, and the wireless
link, and takes no part in either measurement.

\begin{table}[h]
\centering
\caption{The two measurement paths. Neither shares a converter with the other.}
\label{tab:paths}
\small
\setlength{\tabcolsep}{4pt}
\begin{tabular}{@{}p{0.10\textwidth}p{0.24\textwidth}p{0.30\textwidth}p{0.28\textwidth}@{}}
\toprule
Path & Sensing element & Read-out & Reported output \\
\midrule
Pressure & 634 FSR elements at \SI{1}{\centi\metre} pitch
         & Analog multiplexer into the STM32F103RCT6 ADC
         & 645-byte frame (Section~\ref{sec:pressure}) \\[0.4em]
Mass     & Four SC902 load cells
         & Dedicated channel
         & Total load, displayed body mass \\
\bottomrule
\end{tabular}
\end{table}

\paragraph{Compliance and intellectual property.} Table~\ref{tab:compliance}
gives the certification and patent position. It fixes the device as a shipping
product rather than a laboratory rig, which is the condition under which every
service-level number in Section~\ref{sec:perf} was measured.

\begin{table}[h]
\centering
\caption{Certification and intellectual property of the FISICA Scale.}
\label{tab:compliance}
\small
\begin{tabular}{lp{0.63\textwidth}}
\toprule
Area & Standard or reference \\
\midrule
Safety and EMC       & KC, against KS~C~9832:2023 and KS~C~9835:2019 \\
Emissions            & Class B \\
Immunity             & Electrostatic discharge, \SI{\pm 4}{\kilo\volt} contact and \SI{\pm 8}{\kilo\volt} air \\
Conformity           & CE, FCC, and TELEC declarations of conformity; VCCI registration \\
Hazardous substances & RoHS 2011/65/EU with (EU)~2015/863, all ten restricted substances; assessed under IEC~63000:2016+AMD1:2022 with IEC~62321 methods \\
End of life          & WEEE 2012/19/EU with (EU)~2024/884, category~5; rates under IEC/TR~62635:2012 \\
\midrule
Patents registered   & KR~10-1775691, KR~10-1859670 (array read-out and pressure processing) \\
Patents pending      & KR~10-2024-0068497, KR~10-2025-0208869 \\
\bottomrule
\end{tabular}
\end{table}

\begin{table}[t]
\centering
\caption{FISICA Scale hardware specification, from the Care\&Co. product
specification.}
\label{tab:hardware}
\small
\begin{tabular}{p{0.30\textwidth}p{0.60\textwidth}}
\toprule
Item & Specification \\
\midrule
Device size & $321\times321\times55$~mm \\
Pressure sensing & $22\times29$ logical grid at \SI{1}{\centi\metre} pitch,
split into left and right $11\times29$ regions; 634 physical FSR elements \\
Array read-out & Analog multiplexer into the sensor-controller ADC \\
Body-mass sensing & Four SC902 load cells on a separate channel; maximum
specified load 180~kg \\
Sensor controller & STM32F103RCT6 (sensor PCB, FR4) \\
Main controller & ESP32-S3 (main PCB, FR4) \\
Display & 1.3-inch OLED, ENH-0B0130035A \\
Connectivity & 2.4~GHz Wi-Fi and Bluetooth Low Energy \\
Battery & 3.7~V, 1,200~mAh lithium-polymer \\
External power & USB Type-C PD; 5~V/3~A or 9~V/2~A input \\
Top structure & $300\times300\times5$~mm tempered glass \\
Enclosure & ABS case with ABS load-cell brackets and heels on
vibration-damping pads \\
Contact layers & ABS and TPR cover over two silicone layers, one with a
conductive pad \\
Rated power & 5~W \\
\bottomrule
\end{tabular}
\end{table}

\subsection{Deployment}
\label{sec:topology}

Table~\ref{tab:tiers} gives the deployment on which every service-level number
in Section~\ref{sec:perf} was measured. Figure~\ref{fig:topology} in
Appendix~\ref{app:topology} draws the same arrangement.

\begin{table}[h]
\centering
\caption{Deployment tiers. The production and development clusters run the same
stack on different GPUs, which is what makes the comparison in
Section~\ref{sec:mlcost} an internal controlled contrast rather than a vendor
claim.}
\label{tab:tiers}
\small
\setlength{\tabcolsep}{4pt}
\begin{tabular}{@{}p{0.13\textwidth}p{0.19\textwidth}p{0.28\textwidth}p{0.32\textwidth}@{}}
\toprule
Tier & Host & Compute & Role \\
\midrule
Edge        & AWS Seoul, \texttt{t4g.medium}
            & Ubuntu 24.04
            & HTTPS termination (Traefik), central configuration, central
              logging (Grafana, Loki, Promtail) \\[0.4em]
Production  & On-premise, k3s v1.34
            & Control plane, plus $2\times$ (12 cores, \SI{28}{\giga\byte},
              NVIDIA RTX 5060 Ti \SI{16}{\giga\byte})
            & Thirteen microservices at two to three pods each \\[0.4em]
Development & On-premise, k3s, single node
            & 8 cores, \SI{25}{\giga\byte}, NVIDIA RTX 3070 \SI{8}{\giga\byte}
            & Same stack, one pod per service \\[0.4em]
State       & Amazon RDS
            & PostgreSQL
            & Separate production and development databases, per-service
              schema \\
\bottomrule
\end{tabular}
\end{table}

The edge tier reaches the on-premise tier over a WireGuard tunnel. The thirteen
microservices are gateway, authentication, accounts, devices, measurements,
graph and statistics, file storage, notifications, audit, archive, GPU
inference, the conversational agent, and the MCP tool service.

\subsection{Service outputs}

Table~\ref{tab:outputs} separates direct measurements from derived outputs and
recommendations. The split keeps a pressure value from reading as a diagnosis,
and keeps a ranked item from reading as a measured fact.

\begin{table}[t]
\centering
\caption{FISICA service outputs and their inputs.}
\label{tab:outputs}
\small
\begin{tabular}{p{0.17\textwidth}p{0.24\textwidth}p{0.47\textwidth}}
\toprule
Service output & Primary input & Returned information \\
\midrule
Posture analysis & Frontal and lateral RGB images & Body keypoints, depth, a
C7 landmark, and view-specific alignment indicators \\
Pressure analysis & FSR grid and mass channel & Pressure heatmap, left and
right balance, regional loading, foot dimensions, and top classifier labels \\
Embodied view & Fused posture coordinates and the pressure image & An avatar
driven to the measured parameters, with the pressure decal and the
centre-of-gravity line registered to it \\
Footwear & Pressure observations and user constraints & Ranked shoes from the
released catalog, matched features, exclusions, and an explanation trace \\
Exercise & Posture or pressure needs, goals, and safety constraints &
Supported movements, instructions, exclusions, and permitted safety text \\
\bottomrule
\end{tabular}
\end{table}

% ==========================================================================
\section{Plantar-pressure analysis}
\label{sec:pressure}
% ==========================================================================

\subsection{Frame format}

\begin{figure}[t]
\centering
\includegraphics[width=0.34\textwidth]{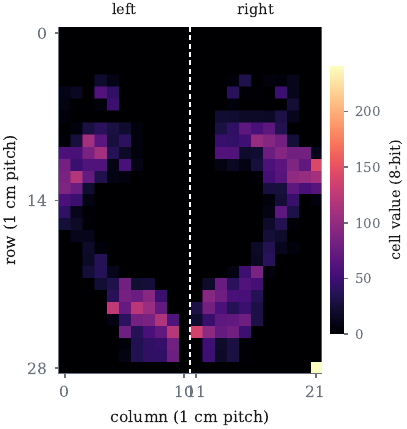}
\caption{A decoded pressure frame in the $22\times29$ product layout. The
dashed line marks the boundary between the two half-mats. The cell pitch is a
physical constant, so foot lengths and widths follow from the grid without a
camera-based scale reference.}
\label{fig:pressure}
\end{figure}

The mat sends a fixed frame of \num{645} bytes while the subject stands still.
Of these, \num{638} bytes form the pressure array as 8-bit values. The product
layout is 22 columns by 29 rows at a \SI{1}{\centi\metre} pitch on both axes.
In row-major software form the same frame has shape $29\times22$. The board
holds \num{634} sensing points, so four logical corner positions have no
sensor. Columns 0 to 10 cover the left foot and columns 11 to 21 the right, so
each foot is sampled by an $11\times29$ region. The remaining bytes carry a
two-byte big-endian body mass in decigrams, a battery field, and two two-byte
terminators that also serve as a frame check. Figure~\ref{fig:pressure} shows
a decoded frame.

\subsection{Foot geometry}

A \SI{1}{\centi\metre} pitch is coarse relative to a foot, and toes often fail
to trip a cell. Foot length is therefore not the extent of the loaded region.
The system estimates it in three steps.

\begin{enumerate}
  \item Compute the principal axis of the loaded cells of each half-mat by
  eigendecomposition of their covariance. Orient the axis heel to toe. Take
  the extent along that axis and dilate it by half a cell diagonal at each end.
  \item Locate the metatarsal ball as the position along the long axis that
  maximizes a product of band width and a prior over the expected location.
  Divide by the anatomical ratio to get a template-completed length.
  \item Report the observed extent, unless the template estimate exceeds it. In
  that case report the template estimate, capped at a fixed multiple of the
  observed extent.
\end{enumerate}

A separate row-profile test decides whether toes registered at all. It uses the
forward gap, the toe-to-ball width ratio, and the presence of a narrowing
between them. The three steps together recover a length in millimetres from a
grid whose pitch is a physical constant, with no camera reference and no user
input.

\subsection{Regional loading}

Regional loading follows the anatomical thirds of the arch-index literature
\cite{cavanagh1987archindex}. The forefoot, midfoot, and rearfoot are the
leading \SI{28}{\percent}, middle \SI{32}{\percent}, and trailing
\SI{40}{\percent} of each foot's bounding rows. The pressure sum in each band
is reported as a percentage of that foot's total. Load symmetry is the ratio of
half-mat sums. A pressure-weighted centroid over the mat is reported,
normalized to the unit square.

This is one static centroid from a single standing frame, not a
centre-of-pressure trajectory over time.

\subsection{Class labels and the report}

A nine-way convolutional classifier reads the same grid as a single-channel
tensor of height 29 and width 22. It has three convolution blocks of
\num{32}, \num{64}, and \num{64} channels with $3\times3$ kernels and pooling
after the first two, followed by two fully connected layers. It emits a
softmax distribution and the top three classes are reported.

\begin{figure}[t]
\centering
\begin{subfigure}[b]{0.23\textwidth}
  \includegraphics[width=\textwidth]{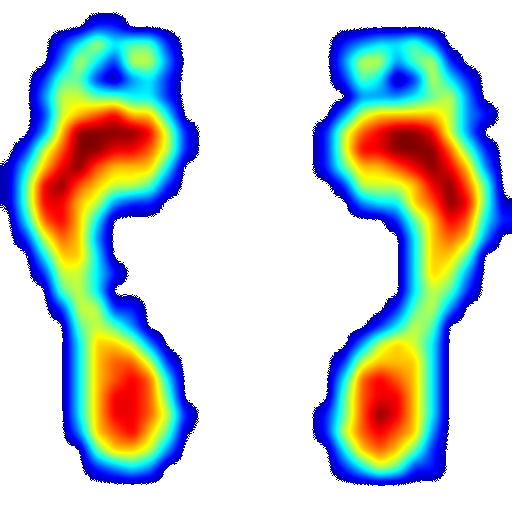}
  \caption{Normal}
\end{subfigure}\hfill
\begin{subfigure}[b]{0.23\textwidth}
  \includegraphics[width=\textwidth]{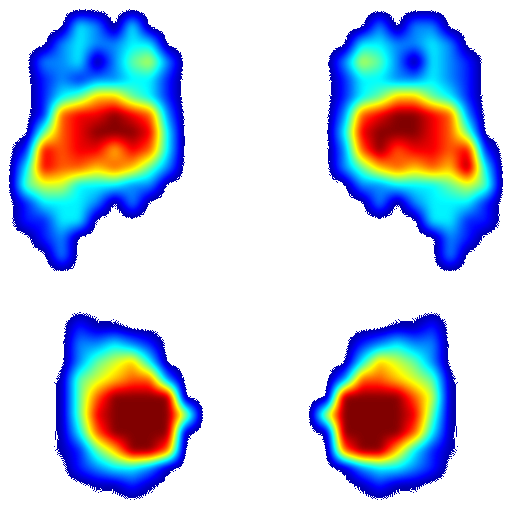}
  \caption{Pes cavus}
\end{subfigure}\hfill
\begin{subfigure}[b]{0.23\textwidth}
  \includegraphics[width=\textwidth]{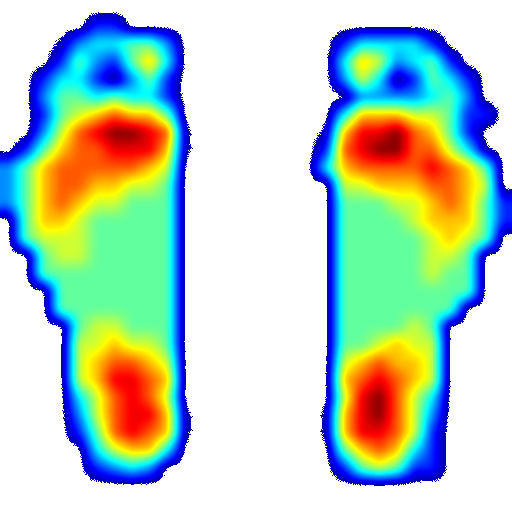}
  \caption{Flat foot}
\end{subfigure}\hfill
\begin{subfigure}[b]{0.23\textwidth}
  \includegraphics[width=\textwidth]{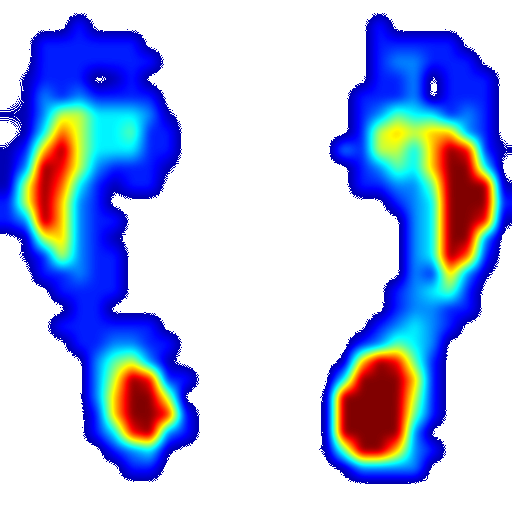}
  \caption{Left-side loading}
\end{subfigure}
\caption{Decoded and interpolated pressure frames for four of the nine class
labels. Each panel shows both feet. The midfoot band is what separates the
first three: a high arch (b) leaves forefoot and heel unconnected, a normal
arch (a) leaves a narrow bridge between them, and a flat foot (c) loads the
midfoot continuously. Panel (d) shows the asymmetry between feet that the
left-right load ratio reports. These are development captures, not class
prototypes, and the label is the classifier output rather than a clinical
finding.}
\label{fig:pressclasses}
\end{figure}

\begin{figure}[t]
\centering
\includegraphics[width=\textwidth]{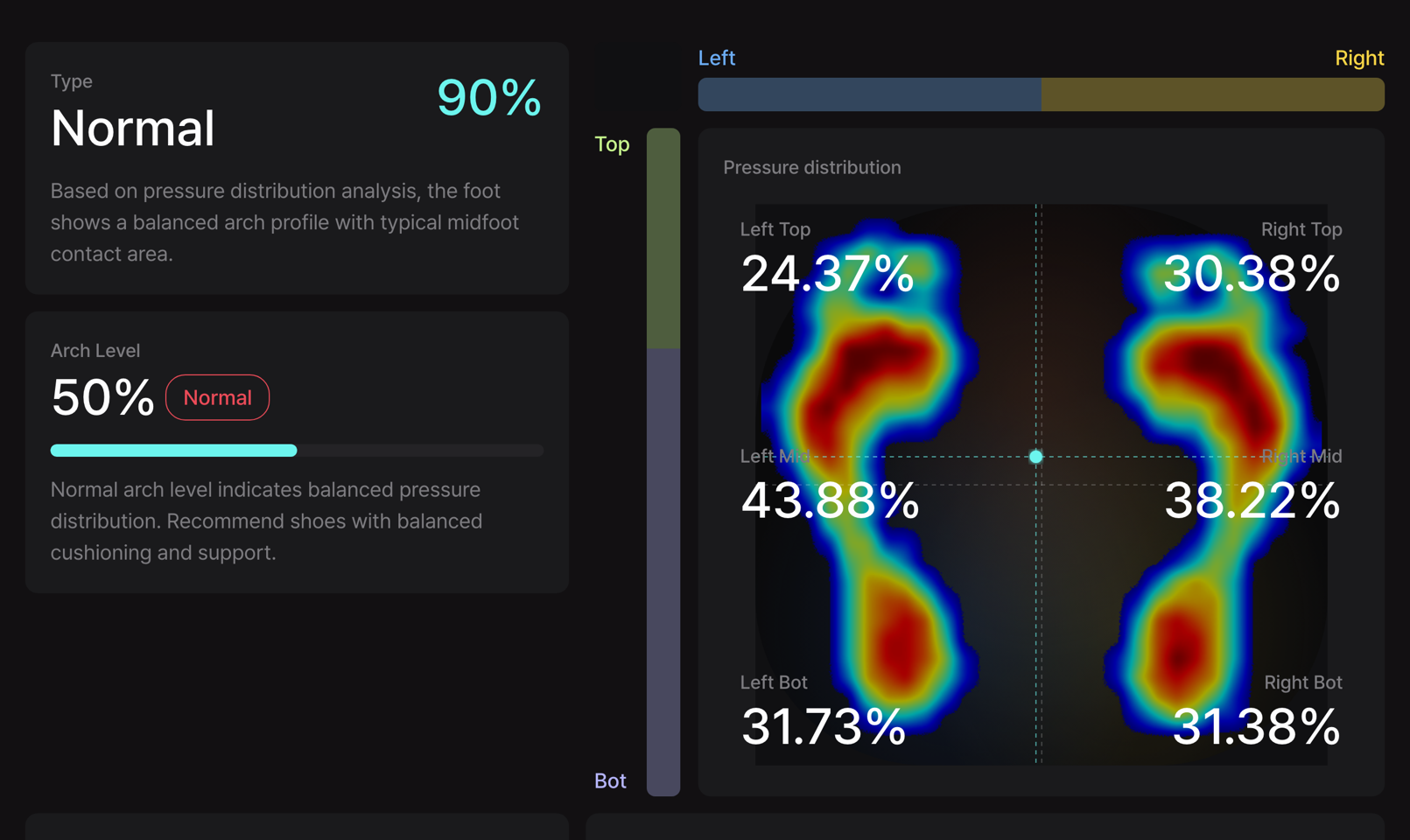}
\caption{Pressure report for the normal example of
Figure~\ref{fig:pressclasses}. The panel gives the top class with its softmax
value, an arch-level index, the left and right load split, and the three
anatomical bands per foot as percentages of that foot's own total. The rest of
the report adds foot length and width in millimetres and a shoe size. The
softmax value is not a calibrated probability and is not validated against a
reference standard.}
\label{fig:pressreport}
\end{figure}

The nine classes are normal, pes cavus, flat foot, lordosis, kyphosis, left
and right scoliosis, and left and right pelvic torsion. This is the same label
set proposed for the reference study in Section~\ref{sec:ethics}.
Figure~\ref{fig:pressclasses} shows decoded frames from four classes.
Figure~\ref{fig:pressreport} shows the report the user sees.

% ==========================================================================
\section{Posture estimation}
\label{sec:vision}
% ==========================================================================

\subsection{Two modes}

The measurement path takes a frontal and a lateral RGB capture and offers two
modes.

The 2D keypoint mode runs an ONNX keypoint-detection graph. It returns joint
coordinates, depth, and a service-specific C7 landmark under measurement schema
1.1.0. Face regions are blurred in the returned imagery. This mode feeds the
geometric extraction and the avatar of Section~\ref{sec:avatar}, and it is the
mode measured in Section~\ref{sec:vision-eval}.

The high-compute mode produces a 3D body reconstruction with mesh-derived
coordinates. Section~\ref{sec:mlcost} reports its service-level cost. Its
accuracy is not measured here.

The measurement service stores the 2D and 3D coordinates, the indicators, the
capture time, the device provenance, and the model response in one
transaction. Only tools that need dense coordinates read them. The C7
coordinate used for cervical posture is stored with the other landmarks and
linked to the image and model response that produced it.

\subsection{Evaluation data}

We measure the deployed 2D keypoint graph on a held-out split of \num{120}
images from the COCO 2017 person-keypoint validation set \cite{lin2014coco}.
It holds \num{168} annotated person instances, and \num{83} of the images
contain one person. We chose this data for three reasons.

It is reproducible. The images, the annotations, and the metric
implementations are public. It is independent of the approved collection in
Section~\ref{sec:ethics}, since full-body photographs and spinal radiographs
identify a person and we release no participant record here. And it does not
match the product operating condition, which is one standing subject filling
the frame at a fixed distance. Section~\ref{sec:angles} handles that mismatch
by reporting error against subject scale instead of averaging over it.

The split does not overlap the \num{1300} images used to fine-tune the
keypoint head. We verified zero image-id overlap. Both came from COCO
\texttt{val2017}, so these values are not comparable to published
\texttt{val2017} results from models that never saw any of it. Comparisons
within Table~\ref{tab:models} hold, since every row uses the same weights and
the same split.

\subsection{Metrics}

\paragraph{Average precision.} We report COCO keypoint AP from
\texttt{pycocotools} with the official per-keypoint sigmas over OKS thresholds
$0.50\!:\!0.05\!:\!0.95$, plus AP$_{50}$ and AP$_{75}$.

\paragraph{PCK.} We state the normalizer, because PCK is not comparable across
papers without it:
\begin{equation}
\mathrm{PCK}@\alpha = \frac{1}{N}\sum_{i=1}^{N}
  \mathbf{1}\!\left[\lVert \hat{p}_i - p_i \rVert_2 < \alpha\, d\right],
  \qquad d = \sqrt{w^2 + h^2},
\label{eq:pck}
\end{equation}
where $w$ and $h$ are the width and height of the ground-truth person bounding
box, and $N$ counts every keypoint annotated as visible. Keypoints of an
instance with no matched prediction count as failures, so detection misses do
not inflate the value.

\paragraph{Instance matching.} Each ground-truth instance with at least one
visible keypoint is matched to the unmatched prediction of highest OKS.
Per-keypoint errors accumulate over matched instances only, and we report the
match rate beside them.

\paragraph{Classification metrics are not reported.} The model emits a fixed
17-keypoint vector per detected instance. There is no true-negative or
false-negative population at the keypoint level, so accuracy, precision, and
recall are not defined for this task.

% ==========================================================================
\section{Measured accuracy and cost of the vision graph}
\label{sec:vision-eval}
% ==========================================================================

\subsection{Setup}

Measurements ran on a workstation with an NVIDIA RTX 4070 SUPER
(\SI{12}{\gibi\byte}), ONNX Runtime \num{1.22} \cite{onnxruntime} with the CUDA
execution provider, CUDA \num{12.6} with cuDNN 9, PyTorch \num{2.11}
\cite{paszke2019pytorch}, and Python \num{3.12} on Windows 11. This is a component bench, not the production
accelerator of Section~\ref{sec:topology}. Section~\ref{sec:perf} gives the
production figures.

All weights are half precision \cite{micikevicius2018mixed} and the runtime
picks convolution algorithms on first use, so every measurement drops \num{20}
warm-up iterations. Device work
is asynchronous, so every stage boundary is synchronized before the clock is
read. Repeating the timed block four times in one process gave median
end-to-end latencies within a factor of \num{1.02}, at a sustained SM clock of
\SI{2775}{\mega\hertz} and a GPU temperature of \SI{48}{\celsius} to
\SI{60}{\celsius}. The measurement is therefore stable enough to support the
comparisons below.

\subsection{Keypoint accuracy}

\begin{figure}[t]
\centering
\includegraphics[width=\textwidth]{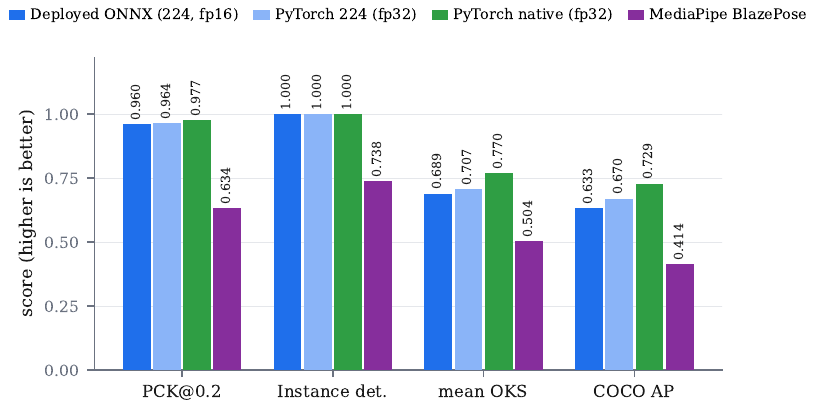}
\caption{Accuracy of the served graph against alternatives on the same split.
The served graph localizes \SI{96.0}{\percent} of visible keypoints within
PCK@0.2 and matches every ground-truth instance. The gap to the on-device
alternative is \num{0.326} PCK and \num{0.219} AP.}
\label{fig:models}
\end{figure}

Table~\ref{tab:models} and Figure~\ref{fig:models} give four configurations on
the same split under the same conditions. Table~\ref{tab:keypoints} gives the
per-keypoint breakdown.

\begin{table}[t]
\centering
\caption{Keypoint estimation on the held-out split of \num{120} public COCO
images and \num{168} person instances. Same images, protocol, and hardware in
every row. PCK@0.2 uses the bounding-box diagonal of
Equation~\eqref{eq:pck}. ``Inst.\ det.''\ is the fraction of ground-truth
instances matched to a prediction. The first three latencies come from the
interleaved run in Table~\ref{tab:ab} and are comparable.
$^{\dagger}$MediaPipe ran separately on a CPU delegate, so its latency is not
comparable to the GPU rows.}
\label{tab:models}
\small
\begin{tabular}{lrrrrrrr}
\toprule
Configuration & AP & AP$_{50}$ & AP$_{75}$ & mOKS & PCK@0.2 & Inst.\ det. & p50 (ms) \\
\midrule
Deployed ONNX ($224^2$, fp16)   & \num{0.633} & \num{0.864} & \num{0.681} & \num{0.689} & \num{0.960} & \num{1.000} & \num{62.8} \\
PyTorch, $224^2$ input (fp32)   & \num{0.670} & \num{0.866} & \num{0.721} & \num{0.707} & \num{0.964} & \num{1.000} & \num{49.4} \\
PyTorch, native res.\ (fp32)    & \num{0.729} & \num{0.857} & \num{0.790} & \num{0.770} & \num{0.977} & \num{1.000} & \num{54.2} \\
MediaPipe BlazePose (heavy)     & \num{0.414} & \num{0.597} & \num{0.457} & \num{0.504} & \num{0.634} & \num{0.738} & \num{134.6}$^{\dagger}$ \\
\bottomrule
\end{tabular}
\end{table}

The deployed configuration matches every ground-truth instance on this split
and reaches PCK@0.2 of \num{0.960} under the normalizer of
Equation~\eqref{eq:pck}.

BlazePose is a single-person, on-device pipeline evaluated here on a
multi-person benchmark, which explains its \num{0.738} instance-detection
rate. On the \num{83} single-person images its AP rises to \num{0.499} and its
PCK@0.2 to \num{0.847}. The deployed graph reaches \num{0.616} and \num{0.953}
on the same subset. We include BlazePose as the alternative a small team would
reach for, not as a head-to-head claim.

\begin{table}[t]
\centering
\caption{Per-keypoint error for the deployed configuration and for
full-resolution inference of the same weights. ``norm.''\ divides the mean
pixel error by the ground-truth bounding-box diagonal. $n$ is the number of
matched visible instances of that keypoint.}
\label{tab:keypoints}
\small
\begin{tabular}{lrrrrrr}
\toprule
& \multicolumn{4}{c}{Deployed ($224\times224$, fp16)} & \multicolumn{2}{c}{Native res.\ (fp32)} \\
\cmidrule(lr){2-5}\cmidrule(lr){6-7}
Keypoint & $n$ & mean (px) & median (px) & norm. & mean (px) & norm. \\
\midrule
\texttt{nose}           & 145 & 7.83  & 2.36  & 0.0260 & 4.23  & 0.0155 \\
\texttt{left\_eye}      & 127 & 4.23  & 2.21  & 0.0196 & 4.64  & 0.0171 \\
\texttt{right\_eye}     & 129 & 7.21  & 1.89  & 0.0236 & 4.38  & 0.0156 \\
\texttt{left\_ear}      & 92  & 5.11  & 3.17  & 0.0161 & 6.31  & 0.0230 \\
\texttt{right\_ear}     & 106 & 9.76  & 3.18  & 0.0326 & 6.21  & 0.0215 \\
\textbf{\texttt{left\_shoulder}}  & 163 & 14.25 & 5.85  & 0.0455 & 10.52 & 0.0290 \\
\textbf{\texttt{right\_shoulder}} & 164 & 10.25 & 5.88  & 0.0351 & 9.42  & 0.0280 \\
\texttt{left\_elbow}    & 144 & 15.65 & 6.15  & 0.0452 & 10.47 & 0.0281 \\
\texttt{right\_elbow}   & 144 & 11.77 & 5.73  & 0.0400 & 9.90  & 0.0269 \\
\texttt{left\_wrist}    & 126 & 18.48 & 5.20  & 0.0470 & 14.03 & 0.0368 \\
\texttt{right\_wrist}   & 138 & 13.06 & 5.64  & 0.0390 & 9.36  & 0.0261 \\
\textbf{\texttt{left\_hip}}       & 147 & 22.74 & 11.12 & 0.0647 & 16.35 & 0.0433 \\
\textbf{\texttt{right\_hip}}      & 147 & 19.41 & 9.50  & 0.0561 & 13.36 & 0.0364 \\
\texttt{left\_knee}     & 96  & 21.56 & 4.63  & 0.0624 & 14.90 & 0.0407 \\
\texttt{right\_knee}    & 92  & 19.75 & 5.33  & 0.0679 & 13.08 & 0.0450 \\
\texttt{left\_ankle}    & 79  & 14.51 & 5.37  & 0.0543 & 10.14 & 0.0389 \\
\texttt{right\_ankle}   & 71  & 10.36 & 4.63  & 0.0417 & 12.80 & 0.0521 \\
\bottomrule
\end{tabular}
\end{table}

Head landmarks localize best, at \num{0.016} to \num{0.033} normalized error,
and shoulders follow at \num{0.035} to \num{0.046}. Hips sit at \num{0.056} to
\num{0.065}, which is a property of the annotation rather than of this model:
the hip joint centre is not visible on the body surface, so annotators infer
it. Section~\ref{sec:angles} therefore treats the pelvic index more
conservatively than the shoulder index.

Table~\ref{tab:ab} gives the latency of the three GPU configurations under one
protocol. We ran all three in one process, interleaved image by image in
rotating order, after warming every path, so clock and thermal drift affect
them equally. The three medians fall within \SI{14}{\milli\second} of each
other, at \SI{62.8}{\milli\second}, \SI{49.4}{\milli\second}, and
\SI{54.2}{\milli\second}. Input resolution and numeric precision therefore move
accuracy without moving the served latency, which is the basis for the export
step in Section~\ref{sec:future}.

\begin{table}[t]
\centering
\caption{Interleaved latency of three configurations sharing weights. One
process, rotating order per image, \num{120} images at \num{3} repetitions
each, after \num{20} warm-up iterations per path. End-to-end, including
preprocessing and transfer of results to host memory.}
\label{tab:ab}
\small
\begin{tabular}{lrrrrr}
\toprule
Configuration & mean (ms) & std (ms) & p50 (ms) & p95 (ms) & p99 (ms) \\
\midrule
Deployed ONNX ($224^2$, fp16) & \num{71.40} & \num{34.22} & \num{62.81} & \num{140.46} & \num{172.70} \\
PyTorch, $224^2$ input (fp32) & \num{53.70} & \num{17.51} & \num{49.44} & \num{88.87}  & \num{109.56} \\
PyTorch, native res.\ (fp32)  & \num{60.65} & \num{19.62} & \num{54.24} & \num{102.05} & \num{136.87} \\
\bottomrule
\end{tabular}
\end{table}

\subsection{Stage cost, dispersion, and memory}
\label{sec:runtime}

Table~\ref{tab:latency} gives stage-separated latency over \num{120} images at
five repetitions each, or \num{600} timed runs.

\begin{table}[t]
\centering
\caption{Stage-separated latency over \num{600} timed runs, after \num{20}
discarded warm-up iterations, with device synchronization at every stage
boundary. Component measurement, not a production service level.}
\label{tab:latency}
\small
\begin{tabular}{lrrrrr}
\toprule
Stage & mean (ms) & std (ms) & p50 (ms) & p95 (ms) & p99 (ms) \\
\midrule
Preprocess    & \num{0.82}  & \num{0.21}  & \num{0.81}  & \num{1.20}   & \num{1.39}  \\
Inference     & \num{81.57} & \num{39.52} & \num{71.40} & \num{158.27} & \num{194.86} \\
Postprocess   & \num{0.12}  & \num{0.03}  & \num{0.11}  & \num{0.17}   & \num{0.21}  \\
\midrule
\textbf{End-to-end} & \num{82.50} & \num{39.56} & \num{72.26} & \num{159.19} & \num{195.85} \\
\bottomrule
\end{tabular}
\end{table}

Three results follow.

\paragraph{Inference dominates.} Preprocessing and postprocessing are
\SI{1.3}{\percent} of the median. A CPU-limited pod is not the bottleneck for
one request, so optimization belongs in the network.

\paragraph{Dispersion comes from instance count.} The p99 is \num{2.7} times
the p50. The exported graph returns a mean of \num{11.6} candidate instances
per image on this split, with a median of \num{7} and a maximum of \num{100}.
Latency tracks that count: images with one instance have a median of
\SI{66.3}{\milli\second}, rising to \SI{119.2}{\milli\second} above twenty
instances (Table~\ref{tab:instcost}). The service keeps only the first
detection, so the keypoint head runs on about a dozen discarded candidates. In
the single-subject condition the distribution sits near
\SI{66}{\milli\second}, and the tail belongs to multi-person evaluation
images, not to deployment.

\begin{table}[t]
\centering
\caption{Inference latency by the number of candidate instances the graph
returns. Per-image median over \num{5} repetitions, \num{120} images.}
\label{tab:instcost}
\small
\begin{tabular}{lrr}
\toprule
Candidate instances & images & median latency (ms) \\
\midrule
$\le 1$    & \num{19} & \num{66.3}  \\
2--5       & \num{29} & \num{65.5}  \\
6--10      & \num{27} & \num{83.6}  \\
11--20     & \num{29} & \num{90.7}  \\
$> 20$     & \num{16} & \num{119.2} \\
\bottomrule
\end{tabular}
\end{table}

\paragraph{Memory splits in two.} Host resident set size peaks at
\SI{1659}{\mebi\byte}, much of it the decoded image set held for the benchmark
rather than the model. Device memory peaks at \SI{1096}{\mebi\byte} above
idle, of which \SI{154}{\mebi\byte} is the half-precision weights at load
time. We report the two separately because one combined figure does not
support capacity planning.

\subsection{Derived angles and subject scale}
\label{sec:angles}

The product displays derived quantities, not keypoints. The shoulder and
pelvic tilt indices come from landmark pairs, and an angle between two points
is more sensitive to localization error than either point.
Table~\ref{tab:angles} gives the absolute error of both indices against
ground-truth landmarks on the same split.

\begin{table}[t]
\centering
\caption{Absolute error of the two derived tilt angles, computed from
predicted and ground-truth landmark pairs on the same instances. ``within
\SI{2}{\degree}'' is the fraction of instances whose derived angle falls within
two degrees of the reference. The mean and p95 follow a small population of
large failures, so the median is the informative statistic.}
\label{tab:angles}
\small
\begin{tabular}{llrrrrr}
\toprule
Configuration & Quantity & $n$ & median & mean & p95 & within \SI{2}{\degree} \\
\midrule
Deployed ONNX ($224^2$, fp16) & shoulder tilt & 159 & 3.16 & 12.40 & 54.06  & 32.1\% \\
Deployed ONNX ($224^2$, fp16) & pelvic tilt   & 144 & 5.26 & 18.62 & 140.59 & 25.0\% \\
PyTorch, $224^2$ (fp32)       & shoulder tilt & 159 & 3.12 & 10.65 & 34.48  & 36.5\% \\
PyTorch, $224^2$ (fp32)       & pelvic tilt   & 145 & 5.86 & 14.23 & 39.50  & 20.7\% \\
PyTorch, native res.\ (fp32)  & shoulder tilt & 159 & 3.16 & 10.21 & 43.71  & 40.3\% \\
PyTorch, native res.\ (fp32)  & pelvic tilt   & 145 & 4.95 & 13.85 & 38.36  & 22.8\% \\
MediaPipe BlazePose (heavy)   & shoulder tilt & 120 & 4.79 & 18.96 & 116.88 & 23.3\% \\
MediaPipe BlazePose (heavy)   & pelvic tilt   & 107 & 5.89 & 23.86 & 148.90 & 17.8\% \\
\bottomrule
\end{tabular}
\end{table}

Two properties of this table matter more than its headline values.

\paragraph{Errors fall into two groups.} Most are small. A few are sign
reversals. For the deployed configuration, \SI{3.8}{\percent} of shoulder-tilt
instances and \SI{6.9}{\percent} of pelvic-tilt instances have the left and
right landmarks swapped, which turns a small angular error into one near
\SI{180}{\degree}. A further \SI{6.9}{\percent} and \SI{9.0}{\percent} are
errors between \num{20} and \SI{90}{\degree}. Excluding both groups, the
median errors are \SI{2.92}{\degree} and \SI{4.03}{\degree}, and the p95 values
fall from \SI{54}{\degree} and \SI{141}{\degree} to \SI{13.5}{\degree} and
\SI{13.3}{\degree}. A tilt index is usually accurate to a few degrees and
sometimes reversed. A service that shows it should detect the reversal and
abstain.

\begin{figure}[t]
\centering
\includegraphics[width=\textwidth]{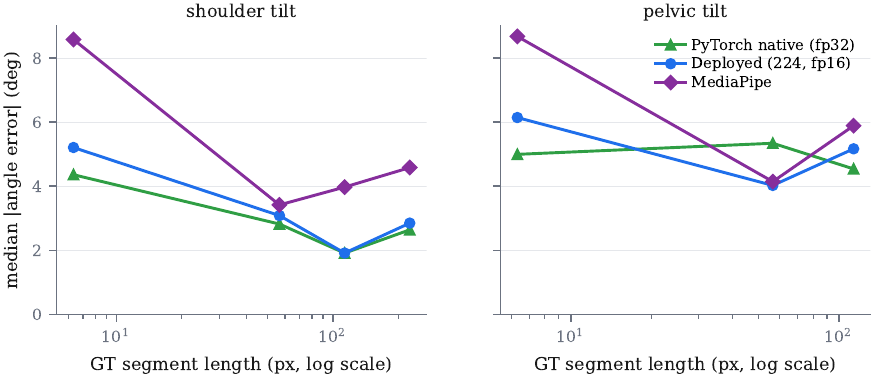}
\caption{Derived-angle error against subject scale. Median error falls and
sign reversals disappear as the landmark pair separates. The pelvic index
stays worse than the shoulder index at every scale, which follows the
per-keypoint hip error in Table~\ref{tab:keypoints}.}
\label{fig:anglescale}
\end{figure}

\paragraph{Error depends on subject scale.} Table~\ref{tab:anglescale} and
Figure~\ref{fig:anglescale} bin the same instances by the pixel separation of
the landmark pair. Below \num{40}~px of shoulder separation the median error is
\SI{5.21}{\degree} and \SI{9.4}{\percent} of instances are reversed. Above
\num{80}~px the median falls to \SI{1.91}{\degree} and no instance is reversed.
The FISICA capture condition, one standing subject filling the frame at a
fixed distance, falls in the largest bin. The marginal values in
Table~\ref{tab:angles} therefore understate the accuracy available in
deployment, because the COCO scale distribution is not the product
distribution. Both tables report COCO subjects and cameras, so the useful
reading is the trend against scale rather than any single bin.

\begin{table}[t]
\centering
\caption{Derived-angle error for the deployed configuration, binned by the
ground-truth pixel separation of the landmark pair. ``reversed'' is the
fraction of instances whose left and right landmarks are swapped.}
\label{tab:anglescale}
\small
\begin{tabular}{llrrrrr}
\toprule
Quantity & Pair separation (px) & $n$ & median (\SI{}{\degree}) & p95 (\SI{}{\degree}) & reversed & within \SI{2}{\degree} \\
\midrule
\multirow{4}{*}{Shoulder tilt}
 & $0$--$40$    & 53 & 5.21 & 147.57 & 9.4\% & 20.8\% \\
 & $40$--$80$   & 58 & 3.09 & 18.44  & 1.7\% & 34.5\% \\
 & $80$--$160$  & 30 & 1.91 & 9.97   & 0.0\% & 50.0\% \\
 & $160$--$320$ & 17 & 2.85 & 33.19  & 0.0\% & 29.4\% \\
\midrule
\multirow{3}{*}{Pelvic tilt}
 & $0$--$40$    & 82 & 6.15 & 157.50 & 9.8\% & 18.3\% \\
 & $40$--$80$   & 39 & 4.03 & 22.61  & 2.6\% & 38.5\% \\
 & $80$--$160$  & 23 & 5.17 & 48.44  & 4.3\% & 26.1\% \\
\bottomrule
\end{tabular}
\end{table}

% ==========================================================================
\section{Embodied posture visualization}
\label{sec:avatar}
% ==========================================================================

The service shows posture as a driven 3D avatar rather than a chart. The
common way to build such a display maps a measured angle onto a rig parameter
through a tuned gain. That makes the avatar an illustration of the finding. We
instead measure the avatar with the same function used on the subject and
solve until the two agree, so the avatar carries the measurement.

The pipeline has three parts. Estimation is the vision service of
Section~\ref{sec:vision}. Extraction is a geometric layer that lifts the
keypoints into an anatomical frame and computes the posture parameters.
Embodiment drives a rigged anatomical avatar until its own measurement
matches the subject.

\subsection{Anatomical frame}

Keypoints are lifted into a body-fixed frame with anterior $-X$, superior
$-Y$, and lateral $+Z$, rather than a camera frame. All angles are then
invariant to capture yaw and to framing. Segment lengths are normalised with
the torso at \num{1.00}, so subject proportions drive avatar limb scaling
independently of stature. A spine polyline runs from root through
$\text{spine}_0$ to $\text{spine}_3$ to neck. Figure~\ref{fig:overlay} shows
the extraction output for two subjects.

\subsection{The sagittal bend metric}
\label{sec:twoline}

Two earlier formulations failed, and each failure shaped the result.

The convexity index takes the maximum posterior deviation of the spine
polyline from its chord, divided by spine length. It depends on sampling.
Because it takes a maximum, a densely sampled curve of 18 to 19 avatar spine
bones always reports a larger value than the five spine points of the subject
for the same shape. Human and avatar cannot be compared, so no loop can close.

The single-joint bend takes the angle at one vertebral node. On real records it
gave \SI{27.7}{\degree} for the normal subject and \SI{28.6}{\degree} for the
kyphotic subject, a separation of \SI{0.9}{\degree}. One joint sees only the
curvature concentrated at that joint, and the apex of a thoracic curve is not
at a fixed anatomical fraction across subjects.

The adopted metric traverses the spine polyline by arc-length fraction. It fits
a lower line between fractions $[a_1,b_1]$ and an upper line between
$[a_2,b_2]$, then returns the angle between them projected onto a plane:
\begin{equation}
\theta = \angle\!\left(
  \mathrm{fit}_{[a_1,b_1]}(\mathcal{S}),\;
  \mathrm{fit}_{[a_2,b_2]}(\mathcal{S})
\right)\Big|_{\Pi}.
\label{eq:twoline}
\end{equation}
The thoracic channel uses $0.25$ to $0.50$ and $0.75$ to $1.00$. The lumbar
channel uses $0.00$ to $0.25$ and $0.25$ to $0.50$. Flexion uses the sagittal
plane and lateral deviation uses the coronal plane.

Three properties follow. The metric is sampling-invariant, because the lines
are set by fraction of spine length, so a five-point human chain and a
nineteen-bone avatar chain return the same value for the same shape. This is
what makes closed-loop matching possible. The metric is apex-agnostic, because
all curvature between the two lines enters the angle. And the metric is
lean-invariant, because it measures an angle between two spine segments, so a
subject who leans forward without changing curvature scores the same. Trunk
lean runs on its own channel, which keeps posture shape and posture
orientation separate.

On the two records that defeated the single-joint formulation, the two-line
thoracic bend gives \SI{42.0}{\degree} and \SI{49.2}{\degree}, a separation of
\SI{7.2}{\degree}. That is about eight times the single-joint separation on the
same data.

Forward-head posture uses the craniovertebral angle (CVA) from the C7 landmark
to the ear in the same frame, where a higher CVA means a more upright neck
\cite{yip2008cva}. On the avatar the C7 vertex sits at the posterior extremity
of the C7 vertebral mesh, the spinous process, rather than at the bone origin,
which lies at the centre of the spine and is not a clinical landmark. The
vertex is anchored to the mesh, so it follows the spine as it flexes.

\subsection{Closed-loop matching}

A secant solver searches each rig parameter over 24 iterations until the
measured avatar value reaches the subject value. The function called on the
avatar is the function used on the human data, and the measured value is passed
as the target with no gain, offset, or rest-pose baseline. The five spinal
channels share one spine and interact, so they are solved in sequence and the
sequence repeats for five rounds. The search runs with the avatar hidden and
only the settled pose is shown.

Four design points came out of failures.

\begin{itemize}
  \item Order matters. The spine is solved before CVA, because thoracic
  flexion moves C7 and changes the neck measurement. Solving CVA first gives a
  stale value. For the same reason the open-loop head push is disabled while
  the loop runs.
  \item Response direction is measured. Before the head channel is solved, the
  rig is probed at its minimum, zero, and maximum forward positions and the
  resulting CVA is logged, because a head handle that drags the cervical bones
  can move the reference point and invert the sign.
  \item Frame lag is a real pitfall. The neck constraint first wrote its result
  only in the engine late-update phase, while the solver measured inside the
  same frame. Every iteration read the previous frame ear position and the
  solver saw almost no response. Exposing the constraint apply step so the
  solver can force it to resolve inside the iteration fixed it.
  \item Saturation is shown. When a channel cannot reach its target within the
  rig range, the badge is marked instead of reporting the clamped value. Rig
  ranges were measured and logged. The lumbar range saturated and was widened
  to $5\times\SI{8}{\degree}=\SI{40}{\degree}$.
\end{itemize}

All displayed values come from this pipeline and the fused coordinates. The
avatar, the badge, the HUD, and the debug overlay read one definition per
channel, with the exceptions listed in Section~\ref{sec:revision}.

\subsection{Rig decoupling and ground registration}

Driving thoracic flexion on a naive skeleton drags the head down and forward
with the spine. The measured CVA is then added on top of the flexion-induced
head displacement and the avatar neck ends far below horizontal. The fix
separates position following from orientation following at the neck. The head
translates with the spine, so the neck stays attached to the trunk, but it does
not inherit the spine rotation. A partial gaze-levelling weight keeps the
result natural. The general point is that each channel must be mechanically
independent on the rig, or the display will exaggerate correlated findings.

At the base of the chain, planted feet plus thoracic flexion make the figure
read as tilted forward. Aligning the heel to the vertical projection of the
pelvis fixes this. The feet, the pressure decal, and the centre-of-gravity line
translate by the same amount, so ground contact stays consistent with the
pelvis and no measured angle changes.

The pressure image is placed as a floor decal at a fixed real-world scale of
\SI{0.29}{\metre} square. It is anchored per foot from the avatar foot-bone
extremes. Foot yaw comes from the heel-to-toe axis of the pressure image, and
anteroposterior stagger comes from the per-foot pressure centroid. The
centre-of-gravity line is drawn from the image centroid in the same placement.
The result is one scene in which sagittal posture, coronal posture, and ground
reaction distribution are registered to each other. This is the reason to fuse
the two modalities.

\subsection{Two-record comparison}

\begin{figure}[t]
\centering
\begin{subfigure}[b]{0.155\textwidth}
  \includegraphics[width=\textwidth]{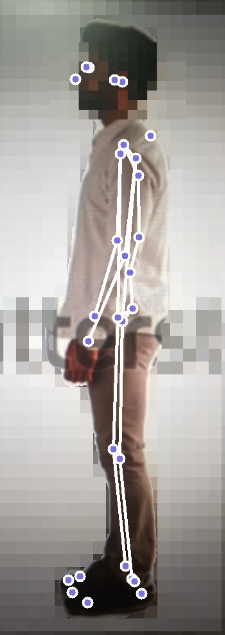}
  \caption{Normal}
\end{subfigure}\hspace{0.06\textwidth}
\begin{subfigure}[b]{0.155\textwidth}
  \includegraphics[width=\textwidth]{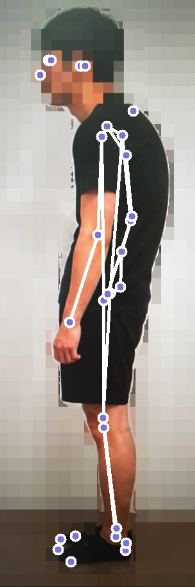}
  \caption{Kyphotic}
\end{subfigure}
\caption{Lateral captures with estimated keypoints. The trunk chain is the
polyline that Equation~\eqref{eq:twoline} uses. The kyphotic subject shows a
posterior convexity of the upper back and a head carried in front of the
shoulder line. Faces are pixelated.}
\label{fig:captures}
\end{figure}

\begin{figure}[t]
\centering
\begin{subfigure}[b]{0.48\textwidth}
  \includegraphics[width=\textwidth]{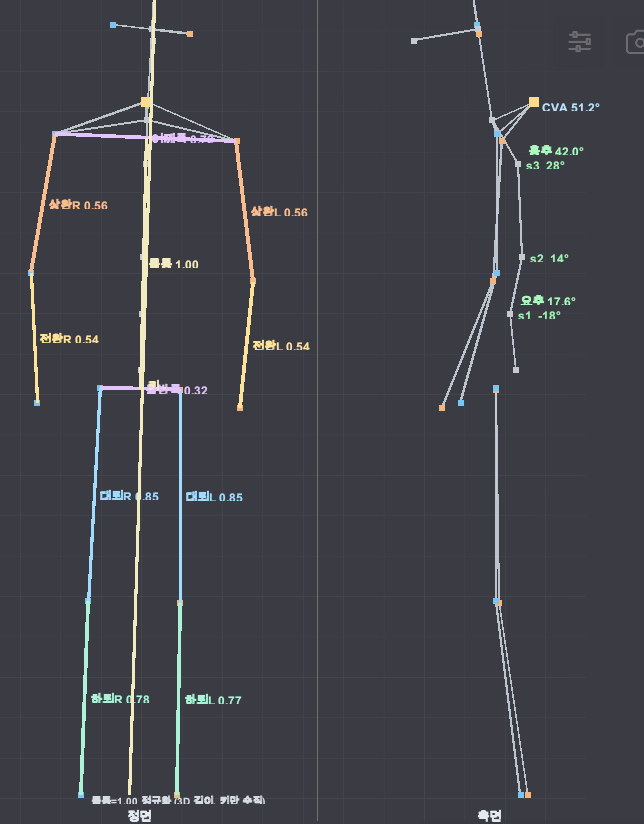}
  \caption{Extraction overlay, normal}
\end{subfigure}\hfill
\begin{subfigure}[b]{0.48\textwidth}
  \includegraphics[width=\textwidth]{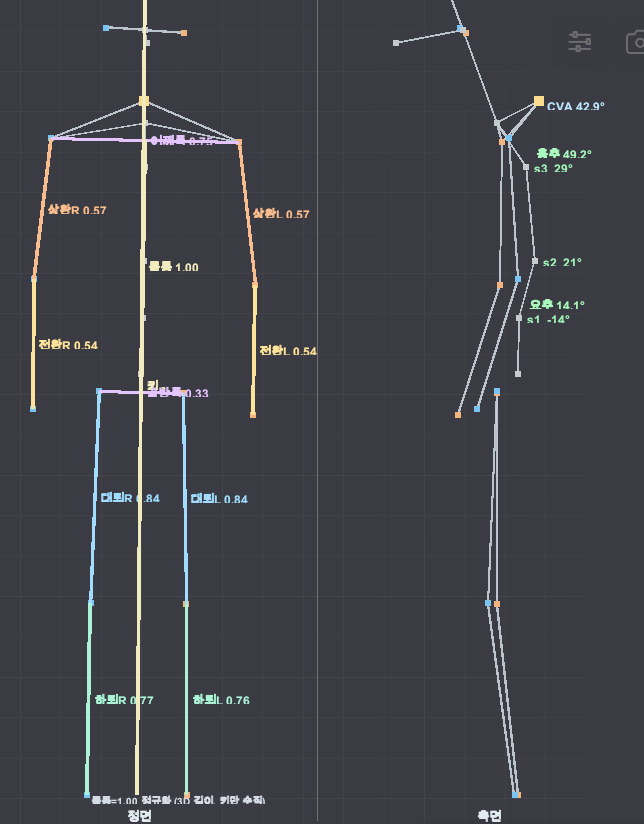}
  \caption{Extraction overlay, kyphotic}
\end{subfigure}
\caption{The extraction overlay. The frontal panel gives torso-normalised
segment proportions. The lateral panel gives the sagittal parameters. In the
kyphotic case the curvature moves upward: the per-node relative flexion $s_2$
rises from \SI{14}{\degree} to \SI{21}{\degree} while $s_1$ moves from
\SI{-18}{\degree} to \SI{-14}{\degree}. A fixed-node metric misses this shape
change. The two-line construction folds it into the \SI{7.2}{\degree}
separation. This developer overlay is compiled out of release builds, so its
labels are the internal Korean identifiers. In the lateral panel they read,
from top, CVA, thoracic bend with node $s_3$, node $s_2$, and lumbar bend with
node $s_1$.}
\label{fig:overlay}
\end{figure}

\begin{figure}[t]
\centering
\begin{subfigure}[b]{0.46\textwidth}
  \centering
  \includegraphics[height=5.3cm]{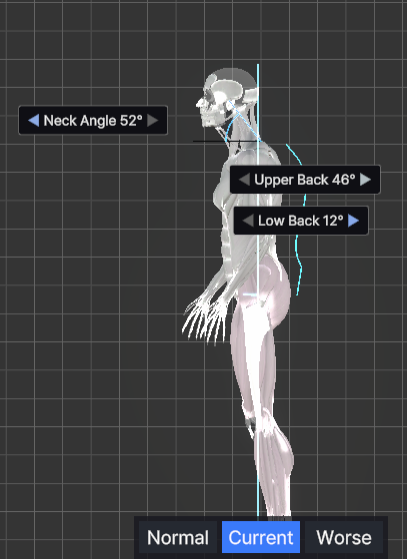}
  \caption{Normal, lateral}
\end{subfigure}\hfill
\begin{subfigure}[b]{0.46\textwidth}
  \centering
  \includegraphics[height=5.3cm]{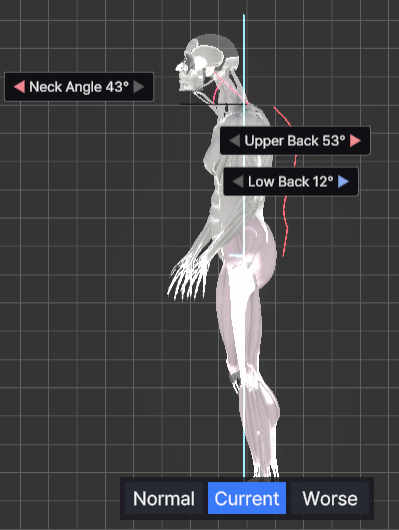}
  \caption{Kyphotic, lateral}
\end{subfigure}

\vspace{1.0em}
\begin{subfigure}[b]{0.46\textwidth}
  \centering
  \includegraphics[height=5.3cm]{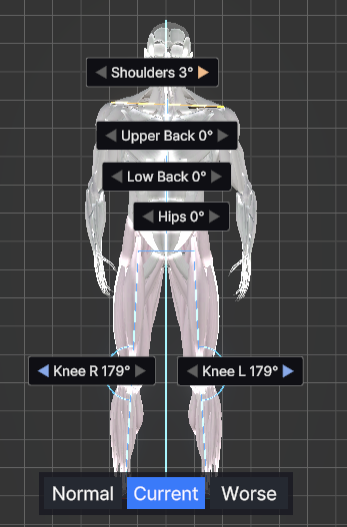}
  \caption{Normal, frontal}
\end{subfigure}\hfill
\begin{subfigure}[b]{0.46\textwidth}
  \centering
  \includegraphics[height=5.3cm]{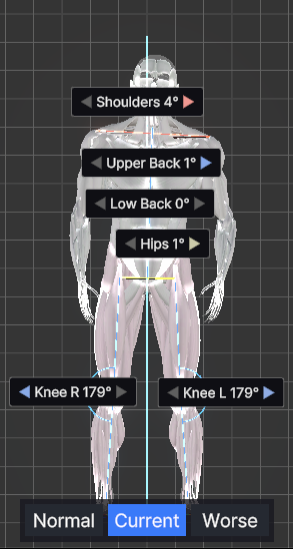}
  \caption{Kyphotic, frontal}
\end{subfigure}
\caption{The driven avatar after the loop converges. Upper Back and Low Back
are measured on the avatar with the functions applied to the subject. The Neck
Angle badge is pinned to the measured value (Section~\ref{sec:revision}). The
spine indicator changes colour above a threshold. Both subjects are near
coronal-plane neutrality in the frontal views, which shows that the sagittal
difference is not an artifact of global misalignment or subject rotation.
Interface labels use plain language, because the reader is the subject.}
\label{fig:avatar}
\end{figure}

Figures~\ref{fig:captures} to~\ref{fig:avatar} and Table~\ref{tab:n2} give a
matched comparison of a normal and a kyphotic record, captured and processed
the same way.

\begin{table}[t]
\centering
\caption{Matched comparison of two records. Rows marked $\dagger$ come from
display paths under revision (Section~\ref{sec:revision}) and are not
independent evidence. The CVA badge is pinned to the measured value each
frame, so its agreement with the subject row is constructed. The shoulder badge
comes from a separate indicator geometry. $n=2$.}
\label{tab:n2}
\small
\begin{tabular}{lrrl}
\toprule
Parameter & Normal & Kyphotic & $\Delta$ \\
\midrule
\textbf{Thoracic two-line bend} (subject)      & \SI{42.0}{\degree} & \SI{49.2}{\degree} & \textbf{\SI{+7.2}{\degree}} \\
Thoracic bend, avatar badge                    & \SI{46}{\degree}   & \SI{53}{\degree}   & \SI{+7}{\degree} \\
\textbf{CVA} (subject)                         & \SI{51.2}{\degree} & \SI{42.9}{\degree} & \textbf{\SI{-8.3}{\degree}}, more forward head \\
CVA, avatar badge$^{\dagger}$                  & \SI{52}{\degree}   & \SI{43}{\degree}   & \SI{-9}{\degree} \\
Lumbar two-line bend (subject)                 & \SI{17.6}{\degree} & \SI{14.1}{\degree} & \SI{-3.5}{\degree} \\
Lumbar bend, avatar badge                      & \SI{12}{\degree}   & \SI{12}{\degree}   & \SI{0}{\degree} \\
Node bend $s_3/s_2/s_1$ (subject)              & \multicolumn{2}{c}{$28/14/{-18}$ \quad$\rightarrow$\quad $29/21/{-14}$} & curvature moves upward \\
Shoulder level, badge$^{\dagger}$              & \SI{3}{\degree}    & \SI{4}{\degree}    & n/a \\
Hip level                                      & \SI{0}{\degree}    & \SI{1}{\degree}    & n/a \\
Knee L / R                                     & \SI{179}{\degree}/\SI{179}{\degree} & \SI{179}{\degree}/\SI{179}{\degree} & n/a \\
\midrule
\emph{(single-joint thoracic bend, rejected)}  & \emph{\SI{27.7}{\degree}} & \emph{\SI{28.6}{\degree}} & \emph{\SI{+0.9}{\degree}} \\
\bottomrule
\end{tabular}
\end{table}

The thoracic channel transfers the difference between the two records almost
exactly, at \SI{7}{\degree} on the avatar against \SI{7.2}{\degree} on the
subject. The avatar reads about \SI{+4}{\degree} above the subject value in
both conditions, so the comparison between conditions holds while the absolute
offset is still under investigation. The lumbar channel shows
\SI{12}{\degree} for both records and is covered in
Section~\ref{sec:revision}.

These are two records used to build and check the metric, not a validation
cohort, and they carry no comparison against a radiographic reference. The
claim they support is that the two-line formulation separates these postures
where the single-joint and convexity formulations do not, and transfers to the
avatar with the difference intact. Section~\ref{sec:ethics} describes the
approved study that supplies the reference standard.

\subsection{Channel status}
\label{sec:revision}

Table~\ref{tab:channels} gives the current state of each channel. The thoracic
and lumbar channels are closed-loop end to end, which is why their solver
residual is visible in Table~\ref{tab:n2}. Three channels are on the roadmap in
Section~\ref{sec:future}. Two of the three put a number on screen, and those
two are marked in Table~\ref{tab:n2} so their rows are read correctly; the
third displays nothing.

\begin{table}[t]
\centering
\caption{Channel status. ``Closed-loop'' means the avatar is measured with the
function used on the subject and solved to match. Rows marked $\dagger$ in
Table~\ref{tab:n2} correspond to the two display paths that are not yet on one
definition.}
\label{tab:channels}
\small
\begin{tabular}{p{0.22\textwidth}p{0.20\textwidth}p{0.20\textwidth}p{0.28\textwidth}}
\toprule
Channel & Drive & Display & Note \\
\midrule
Thoracic bend  & Closed-loop & Solver function & One definition end to end \\
Lumbar bend    & Closed-loop & Solver function & Displays \SI{12}{\degree} on both records; range and saturation marker under review \\
Lateral bend, trunk lean & Closed-loop & Solver function & One definition end to end \\
Craniovertebral angle & Closed-loop & Separate indicator, offset each frame to the measured value & The drive carries the measurement; the badge repeats the input, so it is not convergence evidence \\
Round-shoulder protraction & Tuned gain & Not displayed & No avatar-side measurement; no number reaches the reader \\
Shoulder level & Per-side drop relative to C7, scaled by avatar shoulder width & Separate indicator geometry & Three definitions not yet collapsed to one \\
\bottomrule
\end{tabular}
\end{table}

None of this affects the metric in Section~\ref{sec:twoline}, the
sampling-invariance argument, the \SI{42.0}{\degree} and \SI{49.2}{\degree}
separation, or the thoracic and lumbar badges.

% ==========================================================================
\section{Ontology-grounded recommendation}
\label{sec:recommendation}
% ==========================================================================

The recommendation layer maps structured observations to a fixed catalog of
footwear and exercises. It returns both the candidates and the path that
produced each one. This section describes the \texttt{develop-k3s} state
documented between 18 and 21 August 2026. It is active in the development
environment. It is not evidence of clinical benefit and not a full production
catalog.

\subsection{Knowledge representation}

Domain knowledge lives in PostgreSQL as a lightweight domain ontology rather
than in an RDF/OWL reasoner. Concepts are nodes under stable identifiers.
Meaning between them is stored as a directed edge, written as
$\text{concept}\rightarrow\text{relation type}\rightarrow\text{concept}$. Numeric
thresholds, the required foot type, the rule weight, and whether a rule may
execute are stored in mapping rules. Raw shoe and exercise values are typed
item facts, and functions derived from them are item-feature assertions.
Evidence records and a claim policy bound what a relation is allowed to
explain. User goals, preferences, avoidances, completions, and discomfort are
kept as separate user assertions over the same concept identifiers.

\begin{figure}[t]
\centering
\includegraphics[width=\textwidth]{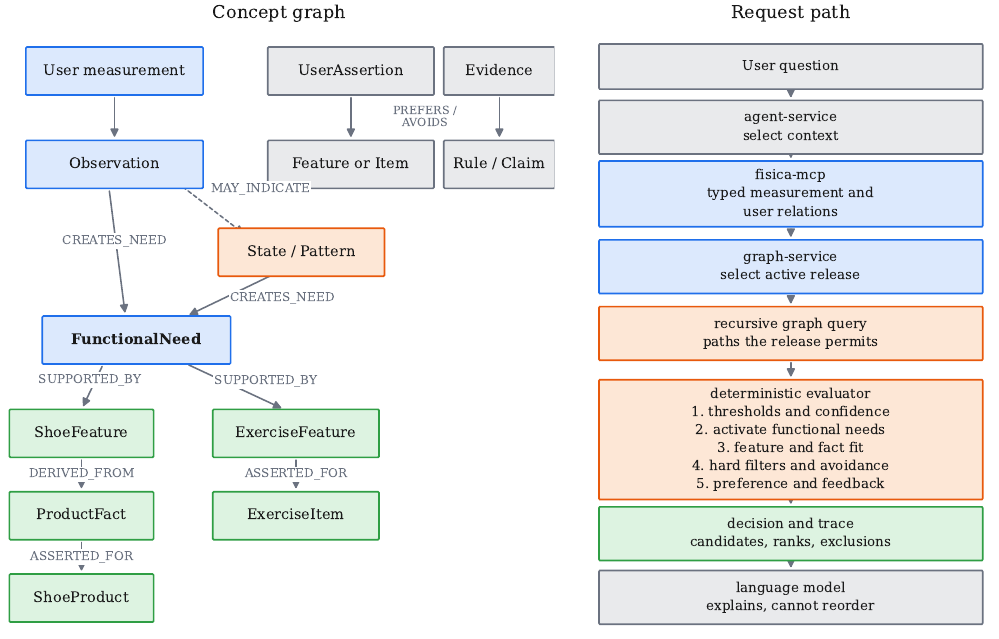}
\caption{Left: the concept graph. An observation creates a functional need
directly, or reaches one through a state or pattern along a
\texttt{MAY\_INDICATE} edge that cannot itself contribute to a score. A need is
supported by shoe and exercise features, which derive from typed item facts.
Right: the request path. Authorization, record scope, and stale or
out-of-range assertions are resolved before the graph is entered, hard
exclusions are applied before any scoring, and the language model receives only
the final candidates and the evidence permitted to explain them.}
\label{fig:ontology}
\end{figure}

Execution has two stages. A PostgreSQL recursive graph query explores the
concept paths the active release permits. A Java evaluator then checks numeric
thresholds, confidence, and composite conditions, applies hard exclusions
before any scoring, and computes the ranked candidates.
Figure~\ref{fig:ontology} shows both the concept graph and the request path,
and Table~\ref{tab:nodes} lists the node and relation inventory. The core path
is
\begin{equation}
 \text{measurement}\rightarrow\text{observation}
 \rightarrow\text{functional need}\rightarrow\text{item feature}
 \rightarrow\text{candidate item}.
 \label{eq:knowledgepath}
\end{equation}

An observation is a computed quantity such as regional pressure or shoulder
alignment. A tendency is not a diagnosis. Relations marked
\texttt{MAY\_INDICATE} or \texttt{SHADOW\_ONLY} can be returned for review but
cannot contribute to an active score. Posture-to-footwear hypotheses about
lordosis, kyphosis, scoliosis, and pelvic rotation exist only as review
material and are excluded from shoe ranking, because the posture channel is
the one whose accuracy depends on scale and can reverse sign
(Section~\ref{sec:angles}).

\begin{table}[t]
\centering
\caption{Node and relation inventory of the domain ontology. The graph is
typed: an evaluator accepts only the need identifiers it supports, so an
arbitrary string cannot enter as a functional need.}
\label{tab:nodes}
\small
\begin{tabular}{p{0.24\textwidth}p{0.32\textwidth}p{0.36\textwidth}}
\toprule
Node & Role & Example \\
\midrule
\texttt{MeasurementRecord} & Source measurement used & record identifier, capture time \\
\texttt{Observation}    & Value computed from the measurement & forefoot load share, shoulder alignment \\
\texttt{StateTendency}  & Tendency the model suggests, not a diagnosis & flat-foot tendency, high-arch tendency \\
\texttt{Pattern}        & Combination of observations or change over time & forward head with round shoulder \\
\texttt{FunctionalNeed} & What an item must support & forefoot pressure redistribution \\
\texttt{ShoeFeature}    & Derived shoe function & heel cushioning, stability support \\
\texttt{ExerciseFeature}& Derived movement function & scapular movement, thoracic mobility \\
\texttt{ProductFact}    & Raw shoe value & mass \SI{240}{\gram}, drop \SI{8}{\milli\metre} \\
\texttt{ExerciseFact}   & Raw movement value & standing, \SIrange{20}{30}{\second} \\
\texttt{RecommendableItem} & Item that may be returned & 699 shoes, 4 exercises \\
\texttt{UserAssertion}  & User relation to a concept or item & brand preference, movement avoidance \\
\texttt{Evidence}       & Source and claim limit of a relation & study, guide, internal review \\
\texttt{MappingRule}    & When a relation may execute & forefoot load $\geq40\%$ \\
\texttt{Release}        & Version of knowledge, catalog, or rules & shoe and exercise releases \\
\texttt{RecommendationEvent} & Snapshot of input and result & candidates, ranks, exclusion reasons \\
\texttt{RecommendationTrace} & Replay information & rule, fact, and evidence identifiers \\
\midrule
\multicolumn{3}{l}{\textbf{Relations}\quad
\texttt{CREATES\_NEED}, \texttt{MAY\_INDICATE}, \texttt{SUPPORTED\_BY},
\texttt{DERIVED\_FROM}, \texttt{ASSERTED\_FOR},} \\
\multicolumn{3}{l}{\quad
\texttt{SUPPORTS\_GOAL}, \texttt{PREFERS}, \texttt{AVOIDS},
\texttt{COMPLETED}, \texttt{EXPERIENCED\_ISSUE},} \\
\multicolumn{3}{l}{\quad
\texttt{SUPPORTED\_BY\_EVIDENCE}, \texttt{GENERATED\_FROM}} \\
\bottomrule
\end{tabular}
\end{table}

Four release types are versioned apart and are immutable once issued:
\texttt{knowledge\_release} for concepts, relations, rules, evidence, and claim
policy; \texttt{catalog\_release} for items, facts, and feature assertions;
\texttt{recommendation\_release} for the combination in use; and an
\texttt{active} pointer per domain. The development environment runs
\texttt{shoe-ontology-2026-08-15.2} with the
\texttt{observation-need-feature-attribute.v2} scorer and
\texttt{exercise-ontology-2026-08-15.2} with the
\texttt{alignment-posture-context-goal-feature.v2} scorer. Before a release
activates, a quality gate checks record counts, identifiers, the meaning of
missing values, a checksum, evidence, and variant coverage.

A recommendation trace records the source measurement and capture time, the
activated observations and context needs, the user assertions and their
confidence, the facts and relation weights, evidence identifiers, hard
exclusions with the reason for each excluded candidate, every release and
scorer version, and a deterministic trace identifier that identifies the same
input. The question ``why was this item recommended'' is therefore answerable
from the data and rules in force at that moment. A bad release rolls back by
moving the active pointer, without rewriting past decisions.

\subsection{Footwear catalog and scoring}

The development catalog holds \num{699} running-shoe models from \num{35}
brands, with \num{10500} typed product facts, \num{7413} feature assertions,
\num{19} mapping rules, and \num{8} evidence records. The laboratory-measured
product attributes are sourced from the RunRepeat shoe-testing database
\cite{runrepeat}, which supplies the weight, drop, stack height, stiffness,
energy return, and traction values that the derived features consume.
Table~\ref{tab:coverage} gives the completeness of each attribute across the
catalog. Missing values stay \texttt{UNKNOWN}. They are not replaced by zero or
a catalog mean, so an item with sparse laboratory data is neither rewarded nor
penalised for the gap. Thirteen derived features are computed from these
facts, including forefoot cushioning, forefoot space, heel cushioning,
stability support, rocker, lightweight, low drop, energy return, soft midsole,
flexibility, traction, breathability, and orthotic accommodation. Each carries
a strength in $[0,1]$, a source-fact coverage value, and the identifier of its
derivation rule, so a strong feature backed by thin data is distinguishable
from a strong feature backed by complete data.

\begin{table}[t]
\centering
\caption{Attribute completeness across the \num{699}-model catalog. Coverage
enters the score directly through $q_f$ in Equation~\eqref{eq:shoe-score},
which is what keeps a sparsely measured item from outranking a fully measured
one on the same nominal strength.}
\label{tab:coverage}
\small
\begin{tabular}{llrr}
\toprule
Attribute & Meaning & Items & Coverage \\
\midrule
\texttt{arch\_support}            & Arch-support class            & 699 & 100\% \\
\texttt{forefoot\_width\_mm}      & Forefoot width at reference size & 699 & 100\% \\
\texttt{removable\_insole}        & Insole detachable             & 699 & 100\% \\
\texttt{orthotic\_friendly}       & Accepts a functional insole   & 699 & 100\% \\
\texttt{weight\_g}                & Mass                          & 688 & 98.43\% \\
\texttt{drop\_lab\_mm}            & Heel-to-forefoot height difference & 674 & 96.42\% \\
\texttt{torsional\_rigidity\_raw} & Torsional rigidity            & 674 & 96.42\% \\
\texttt{heel\_counter\_stiffness\_raw} & Heel-counter stiffness   & 656 & 93.85\% \\
\texttt{breathability\_raw}       & Breathability                 & 615 & 87.98\% \\
\texttt{has\_rocker}              & Rocker construction           & 527 & 75.39\% \\
\texttt{stiffness\_n\_new}        & Bending stiffness             & 437 & 62.52\% \\
\texttt{shock\_absorption\_raw}   & Shock absorption              & 428 & 61.23\% \\
\texttt{energy\_return\_raw}      & Energy return                 & 425 & 60.80\% \\
\texttt{traction\_raw}            & Traction                      & 378 & 54.08\% \\
\texttt{midsole\_softness\_new}   & Midsole hardness              & 242 & 34.62\% \\
\bottomrule
\end{tabular}
\end{table}

Table~\ref{tab:shoe-rules} gives the active measurement-to-need conditions.
These are engineering thresholds used by the development scorer, not validated
clinical cut-points. Foot-type confidence modifies a rule only where stated and
is not used to infer disease.

\begin{table}[t]
\centering
\caption{Active plantar-pressure conditions in the development footwear
scorer. Percentages are fractions of the load on that foot. Confidences are
softmax outputs of the classifier in Section~\ref{sec:pressure}.}
\label{tab:shoe-rules}
\small
\begin{tabular}{p{0.29\textwidth}p{0.29\textwidth}p{0.31\textwidth}}
\toprule
Observation condition & Functional need & Principal shoe feature \\
\midrule
Forefoot load $\geq40\%$ & Forefoot pressure redistribution & Forefoot cushioning \\
Rearfoot load $\geq42\%$ & Heel-impact accommodation & Heel cushioning \\
Midfoot load $\geq30\%$ and flat-foot confidence $\geq0.15$
  & Midfoot support & Stability support \\
Forefoot load $\geq40\%$ and high-arch confidence $\geq0.20$
  & Reinforced redistribution & Cushioning and space \\
Rearfoot load $\geq42\%$ and high-arch confidence $\geq0.20$
  & Reinforced heel accommodation & Heel cushioning \\
\bottomrule
\end{tabular}
\end{table}

For active need $n$ and candidate feature $f$, the measurement contribution is
\begin{equation}
 c_{nf}=s_n\,w_{nf}\,s_f\,q_f,
 \qquad
 S_{\mathrm{base}}=100\,
 \frac{\sum_{n,f}c_{nf}}{\sum_{n,f}s_n w_{nf}},
 \label{eq:shoe-score}
\end{equation}
where $s_n$ is need strength, $w_{nf}$ the released relation weight, $s_f$
feature strength, and $q_f$ source-fact coverage. The evaluator applies a brand
requested in the current turn as a filter, removes avoided brands or products
and recorded product issues, then adds the bounded personalization terms of
Table~\ref{tab:personalization}. It returns three candidates by default, with
ties broken by item identifier so that the same input yields the same order. A
brand named in the current turn overrides a stored preference, while a recorded
product avoidance or fit issue is not cleared by a request for another brand.

\begin{table}[t]
\centering
\caption{Personalization terms, applied after the measurement score and after
hard exclusions. Each is scaled by the confidence of the stored user
assertion. The shoe attribute term is additionally scaled by feature strength
and coverage, so a preference cannot promote an item whose supporting data is
thin.}
\label{tab:personalization}
\small
\begin{tabular}{llp{0.34\textwidth}}
\toprule
Domain & User relation & Effect on score \\
\midrule
\multirow{5}{*}{Footwear}
 & Preferred brand      & $+2.00 \times$ confidence \\
 & Preferred product    & $+1.25 \times$ confidence \\
 & Preferred attribute  & $+0.75 \times$ confidence $\times$ strength $\times$ coverage, capped at $1.5$ per candidate \\
 & Recorded fit issue   & $-100 \times$ confidence \\
\midrule
\multirow{5}{*}{Exercise}
 & Matching goal        & $+0.20 \times$ confidence \\
 & Preferred movement   & $+0.75 \times$ confidence \\
 & Completed movement   & $+0.25 \times$ confidence \\
 & Reported discomfort  & $-1.00 \times$ confidence \\
 & Avoided movement     & Hard exclusion, independent of score \\
\bottomrule
\end{tabular}
\end{table}

\begin{figure}[t]
\centering
\includegraphics[width=0.92\textwidth]{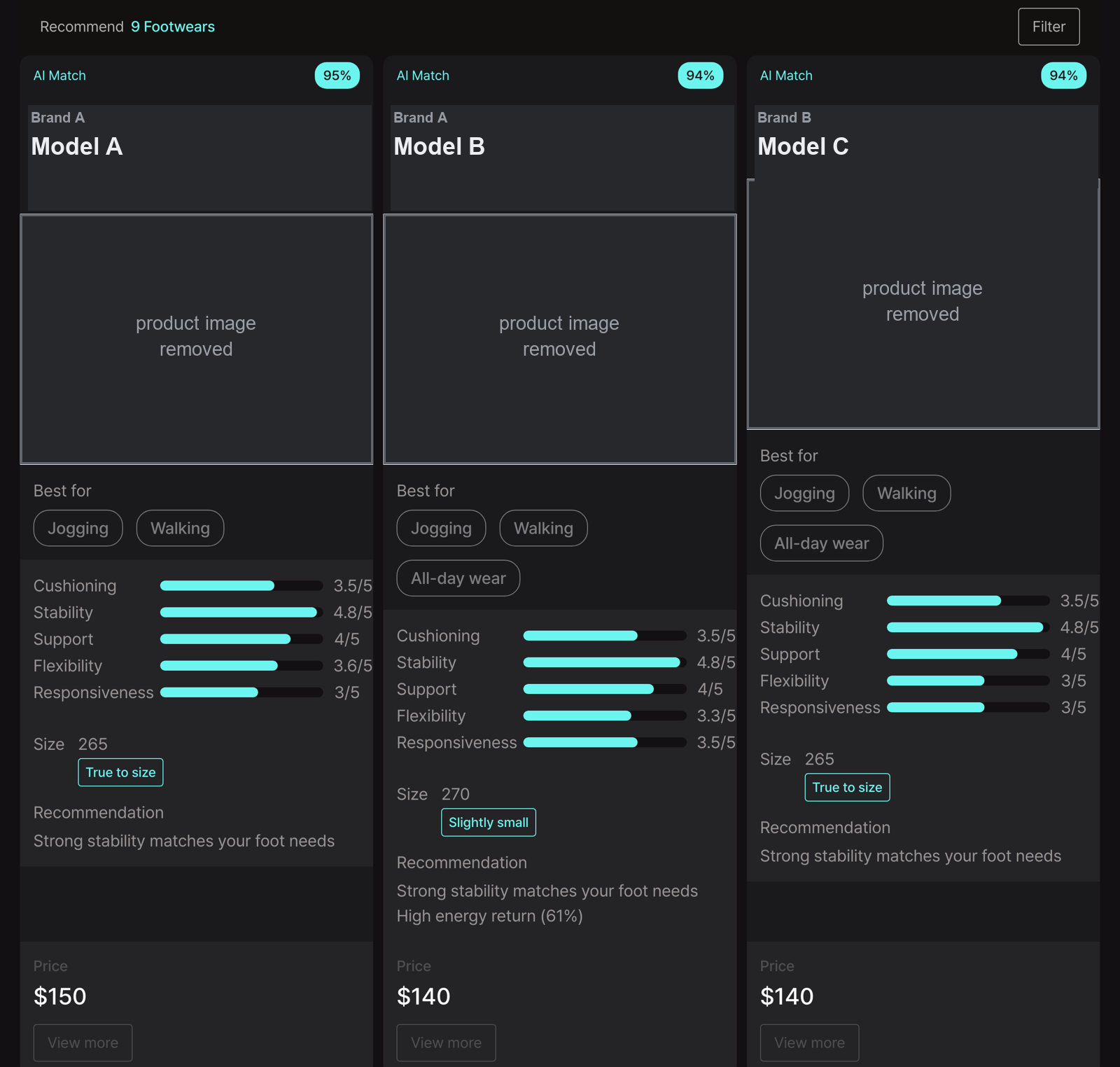}
\caption{Ranked candidates returned for the measurement of
Figure~\ref{fig:pressreport}. The match percentage is a catalog rank from
Equation~\eqref{eq:shoe-score}, not a probability of benefit. The size follows
the measured foot length of Section~\ref{sec:pressure}. Inventory and
size-specific internal dimensions are not in the catalog, so a result is not a
stock or fit guarantee.}
\label{fig:reccandidates}
\end{figure}

Figure~\ref{fig:reccandidates} shows the returned list. Each card carries the
match score of Equation~\eqref{eq:shoe-score}, the per-dimension comparison of
the activated needs against the item feature profile, the size derived from the
measured foot length, and the need that drove the match.
Figure~\ref{fig:itemdetail} in Appendix~\ref{app:topology} gives the full item
page, including the need-against-item radar and the released laboratory
attributes.

\subsection{Exercise catalog and safety}

The exercise release holds four items: W stretch, standing side stretch,
hamstring stretch, and thoracic extension. Three are reachable as
measurement-based candidates through active need or goal relations. Further
movements including thoracic rotation, doorway pectoral stretch, chin tuck,
calf stretch, knee-to-wall, and hip flexor stretch are defined as expansion
candidates. A candidate is not counted in the catalog until it has an item
definition and bilingual labels, facts for region, movement, loading, posture,
equipment and dosage, a relation from a named observation or validated context
need, exclusion and stop conditions, evidence with permitted and prohibited
claims, expert approval, and a new immutable release. Reporting the active
count rather than the candidate list is what keeps the catalog figure and the
reachable figure the same kind of number.

Item facts include target region, movement type, loading, posture, required
equipment, duration or repetitions, bilingual instructions, safety text,
evidence, review status, and permitted and prohibited claims. Rank separates
three contributions:
\begin{equation}
 S_{\mathrm{exercise}}=S_{\mathrm{measurement}}+
 S_{\mathrm{goal}}+S_{\mathrm{feedback}}.
 \label{eq:exercise-score}
\end{equation}
The three terms carry the weights of Table~\ref{tab:personalization}. An
avoidance applies only when the canonical identifier, the canonical name, or
the current display name matches exactly: avoiding the W stretch does not
suppress every shoulder movement. If exclusions remove every supported
candidate, the evaluator returns a conflict or no-result state instead of
inventing a substitute.

Five rules bound the accompanying text. A postural tendency is not stated as a
disease or a confirmed diagnosis. An exercise is not said to correct posture or
treat pain. Muscle shortening, tightness, or cause is not inferred when the
user has not reported it. Only the stop conditions supplied with that movement
are used. A recorded avoidance is not restated as a medical
contraindication. Where the user's own message reports pain, dizziness, or
numbness, the response directs them to a professional.

\subsection{Agent boundary}

The conversational layer exposes \num{16} typed tools across measurement,
posture, footwear, exercise, service information, and user memory. The agent
runs a function-tool loop, and approved calls execute through a separate
MCP-compatible service. The server injects the authenticated user identifier,
resolves the measurement record, merges trusted arguments, enforces a tool
allow-list, rejects duplicate calls, and limits a turn to four tool iterations
and three calls per iteration.

The graph service returns a compact brief with the selected candidates, their
contributions, the evidence allowed for explanation, scope constraints, and
exclusions. The language model turns that brief into the user language. A
response checker rejects unsupported entities, changed rankings, diagnostic or
treatment claims, internal identifiers, and text that presents a user
preference as a measured fact. The language model is an interface to a stored
decision, not the recommendation authority.

% ==========================================================================
\section{Service performance}
\label{sec:perf}
% ==========================================================================

All times are service level, from edge-gateway request receipt to response
completion. They include gateway, microservice, database, and AI processing.
They exclude the Internet round trip between device and server, which adds a
measured \SIrange{0.1}{0.15}{\second} from an external PC.

This is a stricter accounting than most published numbers use. A model-only
figure omits the gateway, the tool call, the database, and the response
assembly. We measure all of it and still land inside the published ranges,
which Section~\ref{sec:baselines} sets out.

\subsection{Method}

Three measurement regimes answer different questions.

\begin{itemize}
  \item Real-usage census. The full edge-gateway access log of the last 30 days
  (2026-07-20 to 08-19) was aggregated in the central log store. ML detail uses
  the 90-day retention window. These are complete sets of user requests, so $n$
  follows call frequency and is small for recently released features.
  \item Recommendation scenario tests. Each feature was called 30 times per
  environment with a test account. Stages were split from the arrival times of
  streaming stage events. A subset of 20 trials was matched against the
  edge-gateway access log processing duration.
  \item Controlled experiment. Fresh accounts were created on 2026-08-24 and
  the measurement-creation API was called 30 times per feature and environment
  with fixed input data. Test accounts were deleted afterwards.
\end{itemize}

Coverage notes: production repeated tests ran in an evening window and an
after-midnight window, set by a per-account daily conversation quota.
Language-model generation varies with external API conditions, and the
after-midnight window carries a latency tail. The 2D and 3D split in
Table~\ref{tab:mlcensus} follows the duration distribution at a two-second
boundary. Confidence intervals are $t$-distribution intervals of the mean at
$\mathrm{df}=29$.

\subsection{Response time by feature}

\begin{figure}[t]
\centering
\includegraphics[width=0.86\textwidth]{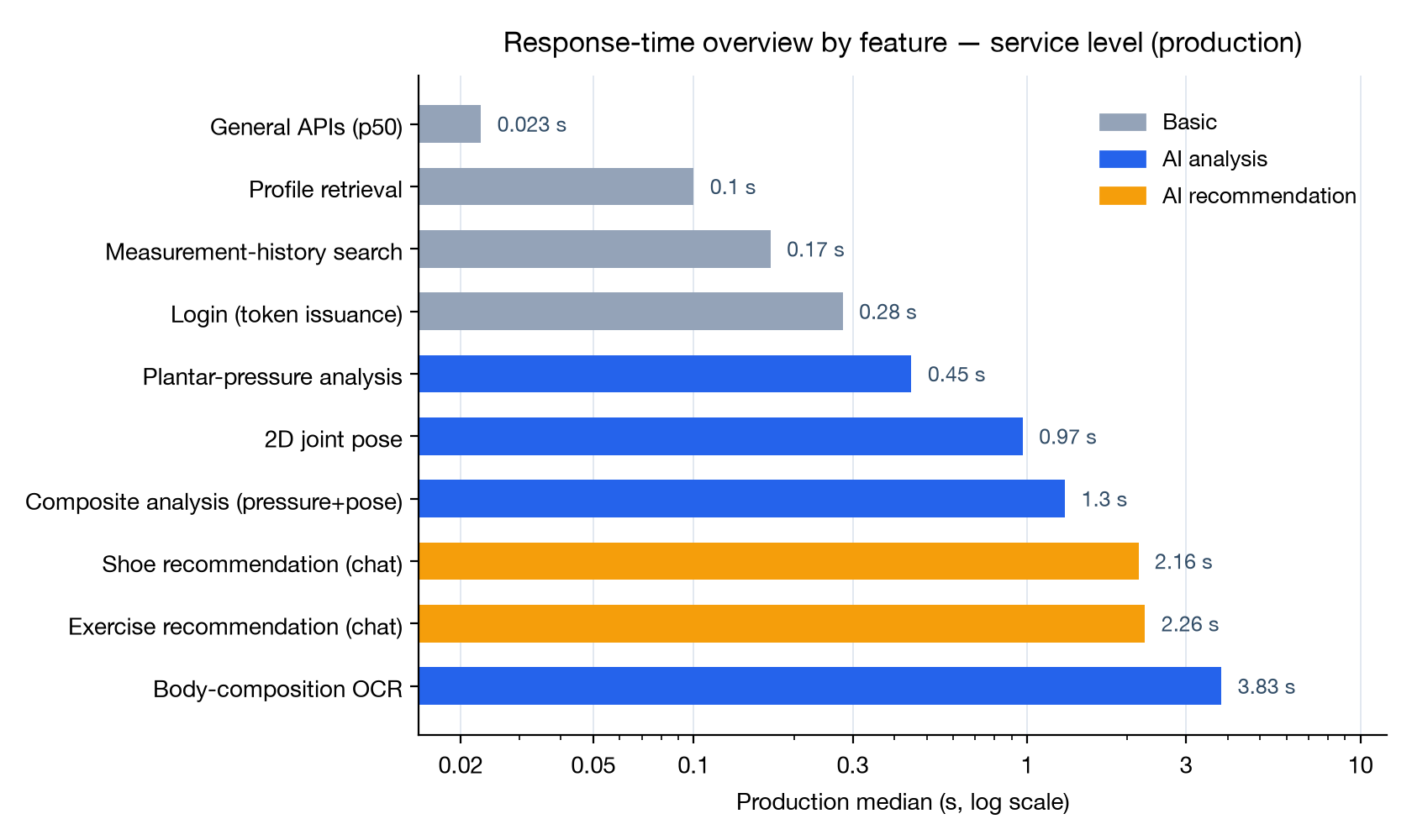}
\caption{Production median response time by feature, service level,
logarithmic axis. Basic read features occupy the left decade, single-model AI
analysis the middle, and agent recommendation the right.}
\label{fig:featureoverview}
\end{figure}

Over the last 30 days of real usage, general APIs had a production median of
\SI{0.023}{\second} and a p95 of \SI{0.249}{\second} over $n=\num{25492}$
requests. The mean is \SI{0.31}{\second}, because a small number of AI and
large-payload requests pull it up. Most requests complete within tens of
milliseconds, so we quote the median for this class. Agent conversations had a
production mean of \SI{2.87}{\second}, a median of \SI{2.51}{\second}, and a
p95 of \SI{6.93}{\second} over $n=104$.
Figure~\ref{fig:featureoverview} places every measured feature on one
logarithmic axis. The spread covers three decades, so one system response time
would not describe this service.

\subsection{Recommendation flow}

\begin{figure}[t]
\centering
\includegraphics[width=\textwidth]{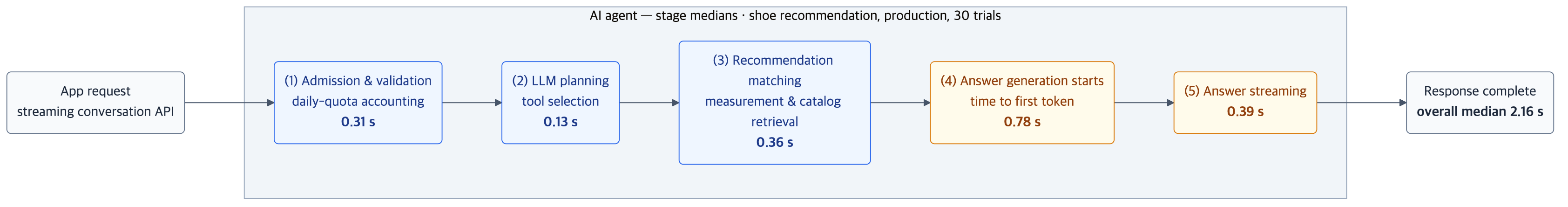}
\caption{Stage decomposition of the shoe-recommendation flow, production, 30
trials, stage medians. Blue stages are rule-based system work. Orange stages
are language-model generation.}
\label{fig:flowstages}
\end{figure}

A shoe recommendation runs through the edge gateway, the API gateway, and the
agent service, then five stages: (1) admission, user validation, and quota
accounting; (2) language-model planning, which selects the tool; (3)
recommendation matching, in which the user pressure and measurement data and
personal memory are retrieved and matched against the shoe catalog by the
evaluator of Section~\ref{sec:recommendation}; (4) language-model answer
generation; (5) answer streaming to completion. The exercise flow is the same,
except that stage 3 retrieves measurement history and condition information
instead of the shoe catalog. Figure~\ref{fig:flowstages} gives the median cost
of each stage.

\begin{table}[t]
\centering
\caption{End-to-end latency at the service level, 30 trials per cell, in
seconds. 95\% CIs are $t$-distribution intervals of the mean at
$\mathrm{df}=29$.}
\label{tab:e2e}
\footnotesize
\setlength{\tabcolsep}{4pt}
\begin{tabular}{llrrlrrr}
\toprule
Feature & Environment & $n$ & Mean $\pm$ SD & 95\% CI & Median & p95 & Min / Max \\
\midrule
Shoe recommendation      & Production  & 30 & $2.45 \pm 1.43$ & [1.92, 2.98] & 2.16 & 3.61  & 1.58 / 9.38  \\
Shoe recommendation      & Development & 30 & $1.93 \pm 0.89$ & [1.60, 2.26] & 1.60 & 3.19  & 1.30 / 5.79  \\
Exercise recommendation  & Production  & 30 & $3.74 \pm 3.55$ & [2.41, 5.07] & 2.26 & 13.26 & 1.74 / 14.47 \\
Exercise recommendation  & Development & 30 & $2.03 \pm 0.14$ & [1.98, 2.08] & 1.99 & 2.27  & 1.84 / 2.41  \\
\bottomrule
\end{tabular}
\end{table}

\begin{table}[t]
\centering
\caption{Stage-level latency, medians in seconds. Stages 1 to 3 are the
rule-based system portion. Stages 4 and 5 are language-model generation.}
\label{tab:stages}
\small
\begin{tabular}{lrrrr}
\toprule
Stage & Shoe prod. & Shoe dev. & Exercise prod. & Exercise dev. \\
\midrule
(1) Admission, validation, quota      & 0.31 & 0.25 & 0.33 & 0.25 \\
(2) Language-model planning           & 0.13 & 0.13 & 0.11 & 0.10 \\
(3) Recommendation matching           & 0.36 & 0.15 & 0.38 & 0.24 \\
(4) Time to first answer token        & 0.78 & 0.73 & 1.07 & 0.62 \\
(5) Answer streaming complete         & 0.39 & 0.36 & 0.31 & 0.76 \\
\midrule
\textbf{System subtotal (1 to 3)}     & \textbf{0.78} & \textbf{0.53} & \textbf{0.84} & \textbf{0.59} \\
\bottomrule
\end{tabular}
\end{table}

\begin{figure}[t]
\centering
\begin{subfigure}[b]{\textwidth}
  \centering
  \includegraphics[width=0.92\textwidth]{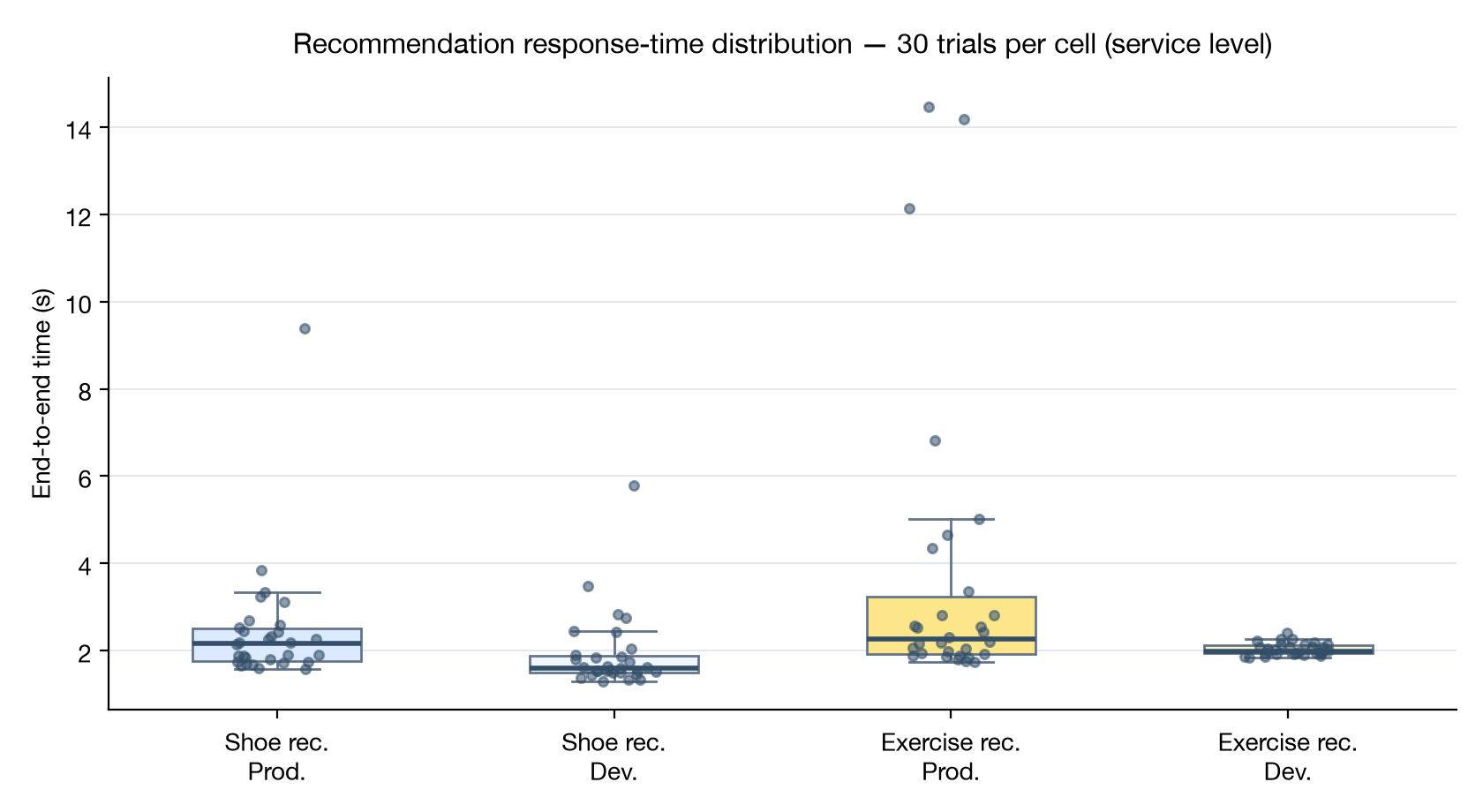}
  \caption{Recommendation response-time distribution, 30 trials per cell}
\end{subfigure}

\vspace{1.0em}
\begin{subfigure}[b]{\textwidth}
  \centering
  \includegraphics[width=0.92\textwidth]{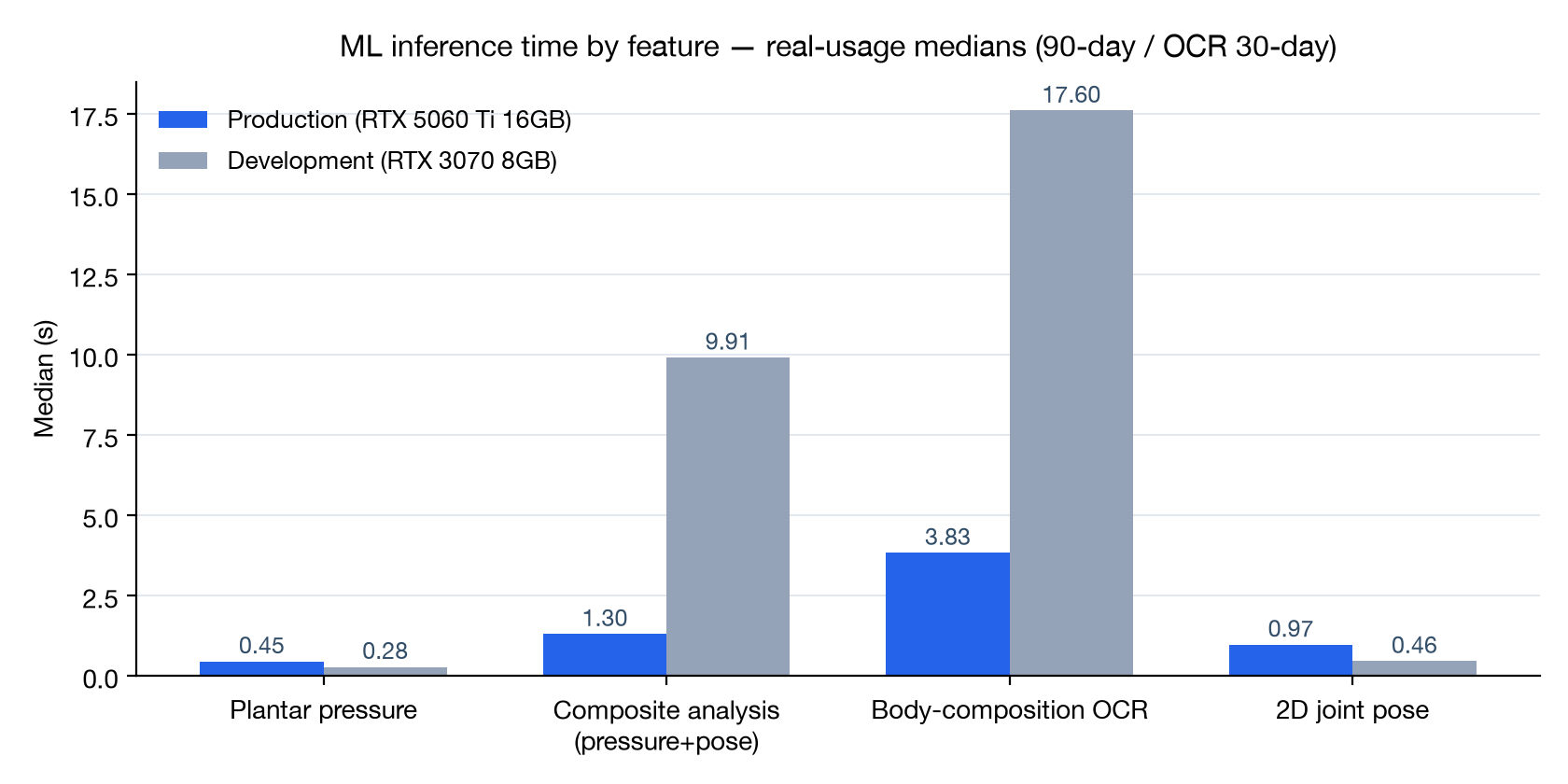}
  \caption{ML inference time by GPU tier, same code in both environments}
\end{subfigure}
\caption{(a) Recommendation response-time distribution. The production tails
are generation-stage outliers, not system-stage outliers. (b) Real-usage
median ML inference time on the production RTX 5060 Ti \SI{16}{\giga\byte}
against the development RTX 3070 \SI{8}{\giga\byte}, same code.}
\label{fig:recdist}
\end{figure}

Tables~\ref{tab:e2e} and~\ref{tab:stages} with Figure~\ref{fig:recdist} give
one result that decides where engineering effort belongs. The rule-based
portion, from request receipt to matching complete, has a production median of
\SI{0.78}{\second} for shoes and \SI{0.84}{\second} for exercises, with small
variance, and stayed under one second in every trial in every environment.
Almost all end-to-end variance comes from the generation stages. The production
exercise flow has an SD of \SI{3.55}{\second} and a p95 of
\SI{13.26}{\second} against a median of \SI{2.26}{\second}. Seven of its 30
trials passed four seconds. One was a first query in a data-empty state, which
took the cold path with stage 3 at \SI{5.1}{\second}. The other six were
generation delays in the after-midnight window. Server capacity is not the
bottleneck, and adding cluster nodes would not move the p95.

\subsection{ML inference in real usage}
\label{sec:mlcost}

Table~\ref{tab:mlcensus} is a census over the 90-day retention window, covering
the ML service from request receipt to analysis completion. Since it is a
census, $n$ follows call frequency. Rows with $n<30$ give the median and range
only, because mean and percentile estimates are unstable at that size.

\begin{table}[t]
\centering
\caption{ML inference statistics, 90-day real-usage census, in seconds.
Production uses an RTX 5060 Ti \SI{16}{\giga\byte}. Development uses an RTX
3070 \SI{8}{\giga\byte}. ``n/a'' marks statistics withheld for $n<30$.}
\label{tab:mlcensus}
\footnotesize
\setlength{\tabcolsep}{4pt}
\begin{tabular}{llrrrrrr}
\toprule
ML feature & Environment & $n$ & Mean & Median & p90 & p95 & Min / Max \\
\midrule
Plantar-pressure analysis      & Production  & 199 & 0.56  & 0.45  & 0.82  & 1.52  & 0.25 / 2.46  \\
Plantar-pressure analysis      & Development & 232 & 0.43  & 0.28  & 0.79  & 1.55  & 0.14 / 2.64  \\
Body-composition report OCR    & Production  & 9   & n/a   & 2.45  & n/a   & n/a   & 0.35 / 4.64  \\
Body-composition report OCR    & Development & 544 & 14.49 & 17.63 & 25.92 & 27.66 & 0.44 / 83.74 \\
Composite measurement analysis & Production  & 351 & 3.10  & 1.30  & 2.19  & 13.17 & 0.77 / 48.82 \\
Composite measurement analysis & Development & 112 & 6.97  & 9.91  & 11.93 & 12.13 & 0.39 / 16.84 \\
Pose analysis, all modes       & Production  & 14  & n/a   & 7.30  & n/a   & n/a   & 0.42 / 46.36 \\
Pose analysis, all modes       & Development & 107 & 2.64  & 0.54  & 10.94 & 11.75 & 0.17 / 15.33 \\
\quad 2D joint pose            & Prod. / Dev.& 5 / 79  & n/a & 0.97 / 0.46 & n/a & n/a & n/a \\
\quad High-compute 3D mode     & Prod. / Dev.& 9 / 28  & n/a & 37.4 / 9.5  & n/a & n/a & max 46.4 / 15.3 \\
\bottomrule
\end{tabular}
\end{table}

Four rows need care, because the direct reading is wrong in each case.

\paragraph{The 3D mode figure is cold loading, not GPU capability.} Models load
lazily at request time, and high-compute calls are rare, at \numrange{174}{822}
hours between bursts. The first call of a burst includes model loading, at
\SIrange{37.4}{46.4}{\second}. Warm calls within minutes take
\SIrange{3.0}{9.5}{\second}, which matches or beats development. Development
looks faster because test traffic kept it warm. A resident model or a warm-up
probe removes this.

\paragraph{The pose row holds two features under one name.} Splitting at the
two-second boundary gives a 2D joint-pose median of \SI{0.97}{\second} in
production and a high-compute median of \SI{37.4}{\second}. The combined
median of \SI{7.30}{\second} describes neither. The request kind should be
logged so the split becomes exact.

\paragraph{OCR gives the one clean GPU comparison.} The same code has a median
of \SI{17.6}{\second} on the development RTX 3070 and \SI{3.83}{\second} on the
production RTX 5060 Ti over the 30-day window, a factor of \num{4.6}. The
pipeline scales with commodity GPU upgrades. The production OCR sample is small
at $n=9$, from a 2026-08-10 rollout, and the 90-day census gives
\SI{2.45}{\second} for the same feature. At that sample size the two windows
are not distinguishable, so the production figure is
\SIrange{2.5}{3.8}{\second}.

\paragraph{Plantar-pressure analysis is the fastest AI path.} At a
\SI{0.45}{\second} production median and a p95 of \SI{1.52}{\second} it sits an
order of magnitude below the recommendation flow, and well inside the one
second that separates an interactive response from a wait
\cite{nielsen1993response}. The plantar-pressure literature reports accuracy
rather than latency, so Table~\ref{tab:baselines} carries no row for it.

\subsection{Controlled experiment}

The census has uncontrolled sample sizes and inputs.
Table~\ref{tab:controlled} fixes both at $n=30$ with identical inputs. These
times cover the full measurement-creation API, including upload and storage, so
they exceed the ML-analysis times of Table~\ref{tab:mlcensus} by
\SIrange{0.3}{1}{\second}.

\begin{table}[t]
\centering
\caption{Controlled measurement experiment on 2026-08-24: fresh accounts,
identical fixed inputs, 30 trials per feature and environment, service level,
in seconds.}
\label{tab:controlled}
\footnotesize
\setlength{\tabcolsep}{4pt}
\begin{tabular}{llrrlrrr}
\toprule
Feature & Environment & $n$ & Mean $\pm$ SD & 95\% CI & Median & p95 & Min / Max \\
\midrule
Measurement, pressure only  & Production  & 30 & $0.61 \pm 0.11$ & [0.57, 0.65] & 0.58 & 0.86 & 0.48 / 0.93  \\
Measurement, pressure only  & Development & 30 & $0.43 \pm 0.03$ & [0.42, 0.44] & 0.43 & 0.49 & 0.38 / 0.50  \\
Measurement, pressure+pose  & Production  & 30 & $3.04 \pm 2.23$ & [2.21, 3.87] & 2.70 & 3.41 & 2.02 / 14.61 \\
Measurement, pressure+pose  & Development & 29 & $2.87 \pm 2.78$ & [1.81, 3.93] & 2.43 & 4.34 & 1.54 / 16.74 \\
\bottomrule
\end{tabular}
\end{table}

One development pressure+pose trial returned an internal error under momentary
contention on the single-GPU node, which leaves $n=29$. The maxima in both
environments, at \num{14.6} and \SI{16.7}{\second}, are tails of the same
contention. The medians and p95 values are stable. The
production-to-development differences are significant under a Mann-Whitney $U$
test, with $p<0.001$ for pressure-only and $p=0.031$ for pressure+pose.
Development is faster, which reflects load rather than GPU tier: the
pressure-only computation is light, so node idleness dominates, while
production carries real traffic and pod routing. Body-composition OCR is
excluded here because it needs a physical report sheet. The census covers it.

\subsection{Comparison with published baselines}
\label{sec:baselines}

Published figures differ in measurement layer, between model-only inference and
service-level end-to-end. Table~\ref{tab:baselines} states the layer in every
row. Our figures are production service-level medians unless stated.

\begin{table}[t]
\centering
\caption{Our measurements against published baselines, with the measurement
layer stated in each row.}
\label{tab:baselines}
\small
\begin{tabular}{p{0.20\textwidth}p{0.26\textwidth}p{0.16\textwidth}p{0.28\textwidth}}
\toprule
Ours (production) & Published baseline & Source & Fairness note \\
\midrule
General REST APIs, p50 \SI{23}{\milli\second}
 & \SI{100}{\milli\second} threshold for instantaneous response
 & \cite{nielsen1993response,google2018rail}
 & Four times of headroom below the classic threshold \\
Agent time to first token, \SIrange{1.59}{1.96}{\second}, including tool orchestration
 & Median TTFT of comparable reasoning-class LLM APIs, \SI{2.92}{\second}, model only
 & \cite{artificialanalysis2026}
 & Ours includes gateway and tool execution; the baseline is a model-only API \\
Agent end-to-end, \SIrange{2.16}{2.26}{\second}
 & Iterative tool-calling agents, \SIrange{11.3}{38.6}{\second} per query, and \SIrange{2.8}{5.9}{\second} after optimization
 & \cite{graphcot2025}
 & Peer-reviewed agent system; the task domain is graph QA \\
Document OCR, \SI{3.83}{\second} per report sheet
 & Commercial cloud OCR APIs over 500 invoices: \SIrange{3}{4}{\second}, \SIrange{4}{5}{\second}, and \SIrange{4}{6}{\second} for three vendors
 & \cite{veryfi2025}
 & Same unit, one document, end-to-end; vendor-run benchmark, so the vendor's own row is the favourable one \\
Composite measurement analysis, \SIrange{2.1}{2.9}{\second}
 & Within the \SI{10}{\second} limit for holding attention on a task
 & \cite{nielsen1993response}
 & Threshold, not a competing system \\
\bottomrule
\end{tabular}
\end{table}

We left out two comparisons. Published pose-model figures are model-forward
only and millisecond scale, such as RTMPose-m at about
\SI{2.3}{\milli\second} per frame on a GTX 1660 Ti \cite{jiang2023rtmpose}, and
MoveNet at \SIrange{30}{100}{FPS} on consumer hardware. Our figure near one
second is a service-level time dominated by image transfer and preprocessing,
so we cite these for context, not for a head-to-head test.
Section~\ref{sec:runtime} holds our model-layer number. And no published
service-latency figure exists for plantar-pressure analysis, so that row is
absent rather than filled with an inapplicable one.

\subsection{Requirements in the product specification}

The device and its software were submitted under a product specification that
sets numeric floors, assessed against ISO/IEC~25023:2016 and
ISO/IEC~25051:2014. Table~\ref{tab:spec} compares those requirements with what
we measure. Both are exceeded by a wide margin: keypoint localization by a
factor of \num{2.4} on the specified metric, and API response time by a factor
of \num{19} at the median.

\begin{table}[h]
\centering
\caption{Product-specification requirements against measured values. PCK uses
the bounding-box diagonal normalizer of Equation~\eqref{eq:pck}, stated so the
comparison can be checked.}
\label{tab:spec}
\small
\begin{tabular}{@{}p{0.26\textwidth}p{0.16\textwidth}p{0.32\textwidth}p{0.18\textwidth}@{}}
\toprule
Requirement & Specified & Measured & Reported in \\
\midrule
Keypoint localization (PCK) & $\geq 40\%$
  & \SI{96.0}{\percent} on the held-out split
  & Section~\ref{sec:vision-eval} \\
API response time & $\leq \SI{450}{\milli\second}$
  & \SI{23}{\milli\second} median, \SI{249}{\milli\second} p95, production
  & Section~\ref{sec:perf} \\
\bottomrule
\end{tabular}
\end{table}

% ==========================================================================
\section{Approved reference-data collection}
\label{sec:ethics}
% ==========================================================================

The reference data this service will be validated against is collected under
institutional review-board approval. Protocol P01-202512-01-047 was reviewed on
29 December 2025, approved, and is valid to 28 December 2026, for the study
\emph{Development of a Deep Learning-Based Model for Predicting Spinal
Alignment Abnormalities Using Posture, X-ray, and Plantar Pressure Data}. The
board classified it at Level 1, minimal risk, and required written consent with
no waiver. Table~\ref{tab:irb} gives the approved terms in full.

The approved collection covers frontal and lateral posture
photographs, frontal and lateral spinal radiographs, and basic information:
participant name, age group, sex, and height. Direct identifiers are not part
of the analysis dataset, and no participant-level result from this protocol
appears in this manuscript.

\begin{table}[t]
\centering
\caption{Approved terms of protocol P01-202512-01-047. The board classified
the study at its lowest risk tier and required written consent, which are the
conditions the data collection now runs under.}
\label{tab:irb}
\small
\begin{tabular}{p{0.30\textwidth}p{0.62\textwidth}}
\toprule
Item & Approved value \\
\midrule
Approval number      & P01-202512-01-047 \\
Review outcome       & Approved, reviewed 29 December 2025 \\
Approval valid to    & 28 December 2026 \\
Study period         & From approval to 31 December 2026 \\
Study type           & Human-subject research, survey research \\
Study form           & Single site \\
Risk level           & Level 1, minimal risk \\
Consent              & Written consent, no waiver \\
Vulnerable subjects  & None included \\
Personal identifiers & Not included in the analysis dataset \\
\bottomrule
\end{tabular}
\end{table}

The planned workflow pairs the frontal posture photograph with the frontal
radiograph, and the lateral photograph with the lateral radiograph, from the
same capture record. Annotators use one surface landmark definition across the
two RGB views and identify the radiographic reference points under a separate
manual. The pairing is at record level. It does not assume that an X-ray pixel
and an RGB pixel share a coordinate system, because camera geometry, field of
view, and projection differ.

\begin{figure}[H]
\centering
\includegraphics[width=\textwidth]{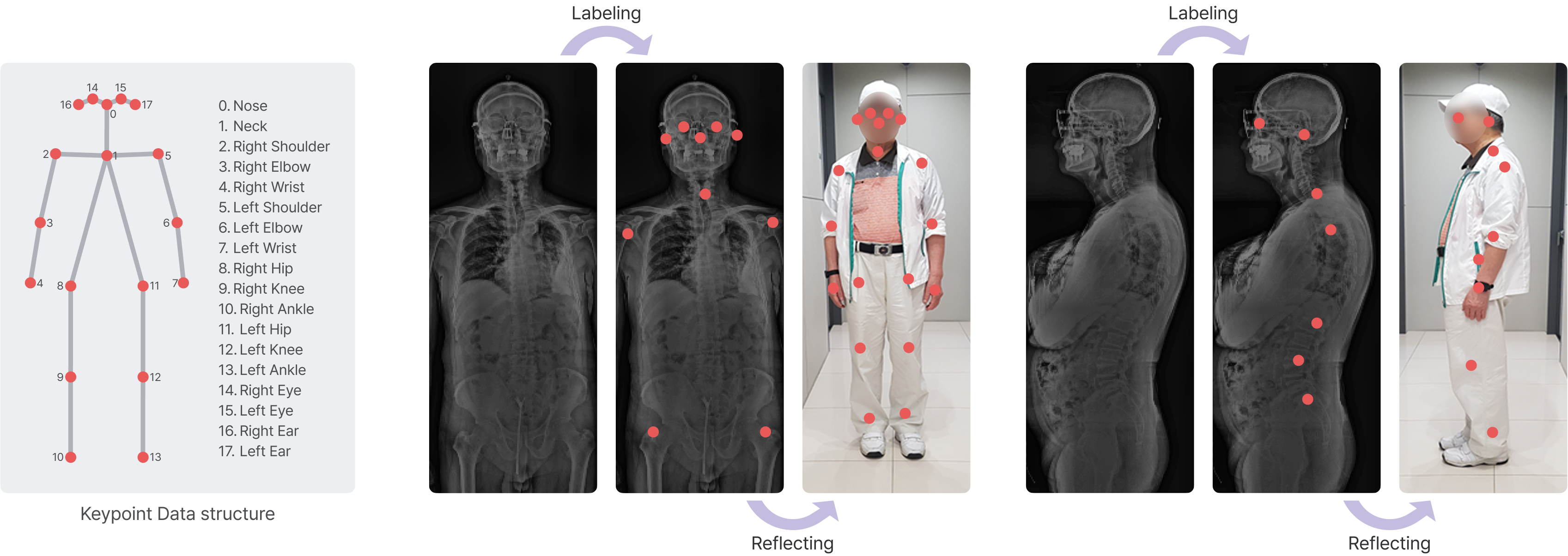}
\caption{Paired-view annotation concept. One surface-landmark schema is shown
with frontal and lateral spinal radiographs and the matching RGB captures. The
diagram records intended landmark correspondence across modalities. It does not
treat their pixel coordinates as interchangeable. Use of this participant image
in a public manuscript remains conditional on confirmation that the consent and
de-identification scope permits publication.}
\label{fig:pairedlabels}
\end{figure}

Figure~\ref{fig:pairedlabels} shows the annotation concept, carried over from
an earlier internal design document for this service. That design listed
candidate labels for normal posture, foot deformity, flat feet, lordosis,
kyphosis, left
and right scoliosis, and left and right pelvic tilt, which is the inventory the
plantar classifier of Section~\ref{sec:pressure} emits. These are proposed
analysis labels, not diagnoses from the current service. The study must define
each label, its reference source, reader qualification, and disagreement
procedure before model validation.

This protocol opens the route to reference-based evaluation: comparison of
image-derived posture coordinates and angles against radiographic landmarks.
The analysis plan covers the final cohort count and exclusions, the interval
and geometry of paired captures, a de-identification procedure, an annotation
manual, blinded readers, inter-reader agreement, and a prespecified statistical
test. This is an approved data resource for prospective validation.

% ==========================================================================
\section{Conclusion}
% ==========================================================================

FISICA runs as one service, not a set of separate models. A pressure frame and
two photographs become a versioned measurement record, foot and posture
findings, a driven avatar, and a report with shoe and exercise candidates.
Rules control filtering and ranking. The language model selects approved tools
and explains stored results.

The measured position is as follows. The rule-based portion of every
recommendation completes in under one second, and it held under one second in
every trial in both environments. Plantar-pressure analysis returns in
\SI{0.45}{\second} and general APIs in \SI{0.023}{\second}. Agent
recommendation returns in \SI{2.16}{\second} for shoes and \SI{2.26}{\second}
for exercises, against published tool-calling agent figures of
\SIrange{2.8}{5.9}{\second} after optimization. The deployed keypoint graph
reaches \num{0.633} AP and \num{0.960} PCK@0.2 on public data and matches every
ground-truth instance on the evaluation split. The same pipeline scales with
commodity GPUs, at a factor of \num{4.6} between the two GPU tiers we run. The
catalog holds \num{699} shoes with \num{10500} typed facts, and every ranking
carries a trace back to the measurement that produced it.

% ==========================================================================
\section{Future work}
\label{sec:future}
% ==========================================================================

\paragraph{Vision export.} Move the served graph to native resolution in full
precision. Table~\ref{tab:models} gives the accuracy this makes available,
\num{0.097}~AP and \num{0.081} mean OKS, and Table~\ref{tab:ab} shows the three
configurations sitting within \SI{14}{\milli\second} of each other, so the
change fits the current latency budget.

\paragraph{Abstention on derived angles.} Section~\ref{sec:angles} shows that
tilt-index errors separate into small errors and left-right reversals, and that
reversals disappear above \num{80}~px of landmark separation. A scale gate plus
a reversal check lets the service withhold the index instead of displaying it,
which converts the failure mode into an abstention.

\paragraph{Avatar channels.} Bring craniovertebral angle, round-shoulder
protraction, and shoulder level onto one definition end to end, as the thoracic
and lumbar channels already are (Table~\ref{tab:channels}). Read the
per-channel solver residual to resolve the thoracic offset and confirm the
lumbar range and saturation marker.

\paragraph{Instrumentation.} Log the pose request mode so the 2D and
high-compute figures separate exactly, and keep the high-compute model resident
or warmed so first-call loading leaves the served latency
(Section~\ref{sec:mlcost}).

\paragraph{Catalog depth.} Add inventory state and size-specific internal
dimensions, and activate the relation for the remaining exercise item
(Section~\ref{sec:recommendation}).

\paragraph{Reference validation.} Execute protocol P01-202512-01-047 to
quantify agreement between image-derived posture angles and radiographic
landmarks (Section~\ref{sec:ethics}).

\bibliographystyle{plainnat}
\bibliography{refs}

\appendix

% ==========================================================================
\section{Supplementary figures}
\label{app:topology}
% ==========================================================================

\begin{figure}[t]
\centering
\includegraphics[width=\textwidth]{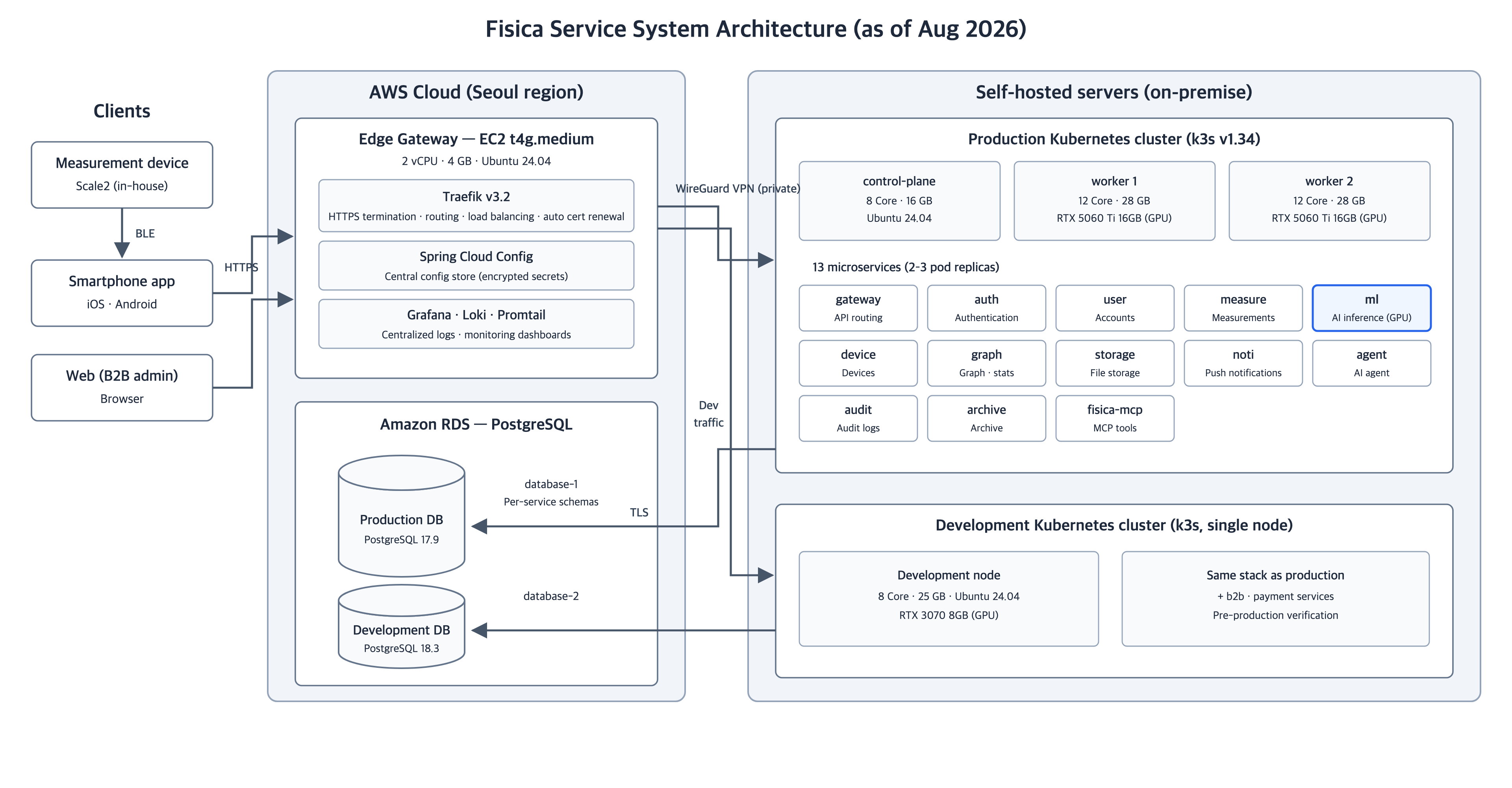}
\caption{Deployed topology. An AWS edge tier handles HTTPS termination,
central configuration, central logging, and managed PostgreSQL. A WireGuard
tunnel connects it to the on-premise k3s clusters: a production cluster of two
GPU worker nodes running thirteen microservices at two to three pods each, and
a single-node development cluster running the same stack.}
\label{fig:topology}
\end{figure}

\begin{figure}[p]
\centering
\includegraphics[height=0.90\textheight]{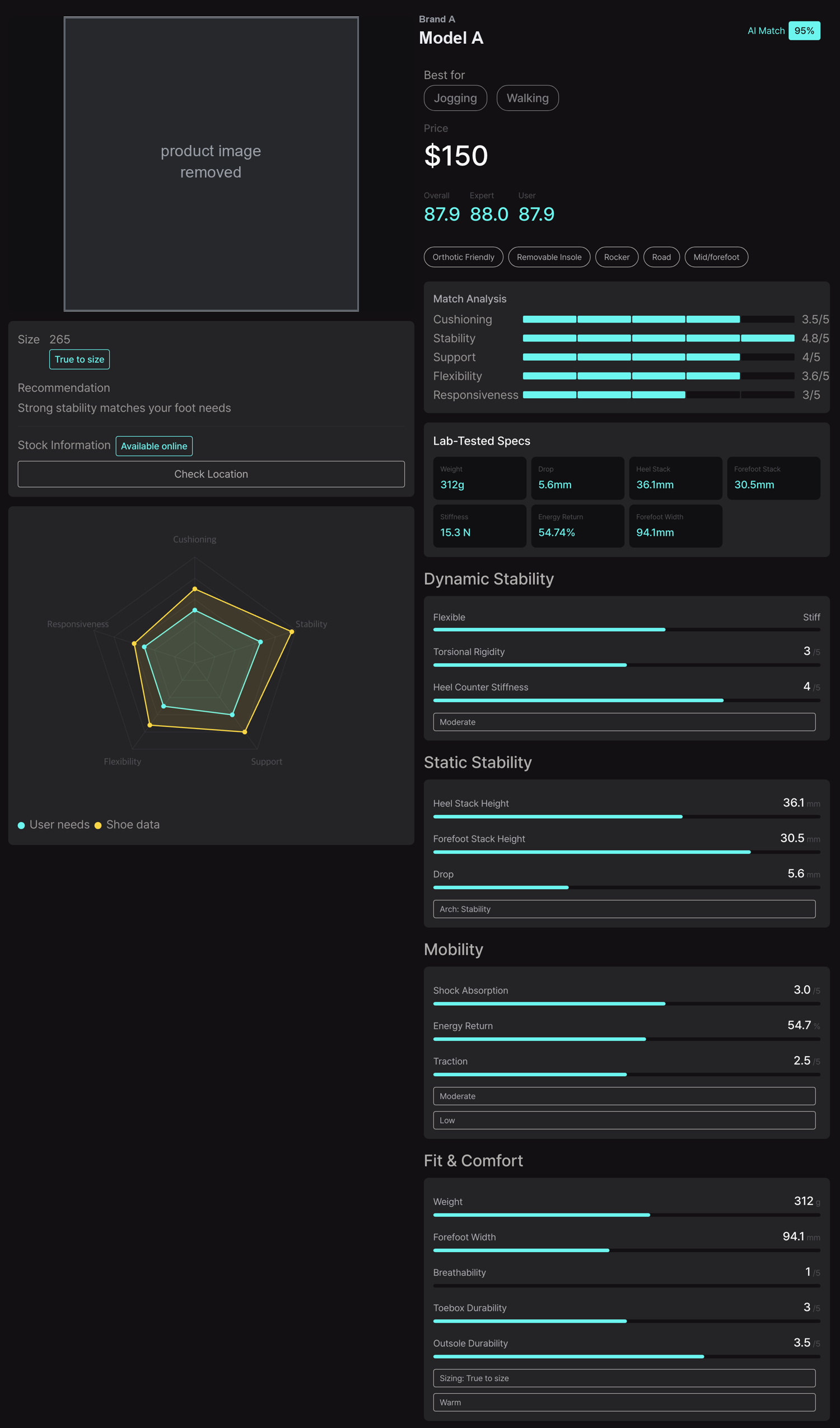}
\caption{Item page for the top-ranked candidate. Left: fit, size, stock state,
and the need-against-item radar, where the inner outline is the user need
profile and the outer outline is the item profile. Right: the match score, the
per-dimension breakdown, and the released facts grouped into dynamic
stability, static stability, mobility, and fit. Every number on this page
traces to a catalog fact identifier and a derivation rule.}
\label{fig:itemdetail}
\end{figure}

Two figures are reproduced at full size here rather than in the body.
Figure~\ref{fig:topology} draws the deployment described in
Section~\ref{sec:topology}, on which every service-level time in
Section~\ref{sec:perf} was measured.
Figure~\ref{fig:itemdetail} gives one candidate page from the same measurement
as Figure~\ref{fig:pressreport} and Figure~\ref{fig:reccandidates}, and shows
the chain end to end: the activated need profile against the item feature
profile, the match score, and the released laboratory attributes that produced
each feature strength in Equation~\eqref{eq:shoe-score}.

\end{document}